\documentclass[twoside,journal]{IEEEtran}

\usepackage{color}
\usepackage{cite}
\usepackage{amsmath}
\usepackage{amssymb}
\usepackage{graphicx}
\usepackage{epstopdf}
\usepackage{tabularx}
\usepackage{booktabs} 
\usepackage[usenames,dvipsnames]{xcolor}
\usepackage{eso-pic} 
\usepackage{textcomp} 
\usepackage{pifont}
\usepackage{anyfontsize}
\usepackage{changepage}
\usepackage{setspace}
\usepackage{url}
\usepackage{multirow}
\usepackage{arydshln}

\ifCLASSINFOpdf
\else
\fi

\AddToShipoutPicture*{\small \raisebox{0.65cm}
{\hspace*{2.2cm}
\begin{minipage}[c]{0.95\textwidth}
\footnotesize
{\setlength{\baselineskip}{0.5\baselineskip}
\textcopyright 2019 IEEE. Personal use of this material is permitted.  Permission from IEEE must be obtained for all other uses, in any current or future media, including reprinting/republishing this material for advertising or promotional purposes, creating new collective works, for resale or redistribution to servers or lists, or reuse of any copyrighted component of this work in other works.
\par}
\end{minipage}}}%

\AddToShipoutPicture*{\small \raisebox{1.2cm}
{\hspace*{-17.35cm}
\begin{minipage}[c]{0.95\textwidth}
\footnotesize
{\setlength{\baselineskip}{0.5\baselineskip}
This document is the paper as accepted for publication in TBioCAS, the fully-edited paper is available at \url{https://ieeexplore.ieee.org/document/8764001}.
\par}
\end{minipage}}}%

\begin{document}

\title{\vspace*{-3.5mm}MorphIC: A 65-nm 738k-Synapse/mm$^2$ Quad-Core Binary-Weight Digital Neuromorphic Processor\\with Stochastic Spike-Driven Online Learning}
\author{Charlotte~Frenkel,~\IEEEmembership{Student Member,~IEEE,}
        Jean-Didier~Legat,~\IEEEmembership{Senior Member,~IEEE,}\\
        and~David~Bol,~\IEEEmembership{Senior Member,~IEEE}\vspace*{-3.5mm}%
\thanks{Manuscript received xxxxx xx, 2019; revised xxxxx xx, 2019; accepted xxxxx xx, 2019. Date of publication xxxxx xx, 2019; date of current version xxxxx xx, 2019.}%
\thanks{The authors are with the ICTEAM Institute, Universit\'e catholique de Louvain, Louvain-la-Neuve BE-1348, Belgium (e-mail: \{charlotte.frenkel, jean-didier.legat, david.bol\}@uclouvain.be).}%
\thanks{C. Frenkel is with Universit\'e catholique de Louvain as a Research Fellow from the National Foundation for Scientific Research (FNRS) of Belgium.}%
\thanks{Color versions of one or more of the figures in this paper are available online at http://ieeexplore.ieee.org.}%
\thanks{Digital Object Identifier 10.1109/TBCAS.2019.2928793}}%

\markboth{IEEE TRANSACTIONS ON BIOMEDICAL CIRCUITS AND SYSTEMS,~VOL.~xx, NO.~xx, XXXXX~2019}%
{FRENKEL \MakeLowercase{\textit{et al.}}: A Quad-Core Binary-Weight Processor with Stochastic Spike-Driven Online Learning}%

\maketitle

\begin{abstract} 
Recent trends in the field of neural network accelerators investigate weight quantization as a means to increase the resource- and power-efficiency of hardware devices. As full on-chip weight storage is necessary to avoid the high energy cost of off-chip memory accesses, memory reduction requirements for weight storage pushed toward the use of binary weights, which were demonstrated to have a limited accuracy reduction on many applications when quantization-aware training techniques are used. In parallel, spiking neural network (SNN) architectures are explored to further reduce power when processing sparse event-based data streams, while on-chip spike-based online learning appears as a key feature for applications constrained in power and resources during the training phase. However, designing power- and area-efficient spiking   neural networks still requires the development of specific techniques  in order to leverage on-chip online learning on binary weights without compromising the synapse density. In this work, we demonstrate MorphIC, a quad-core binary-weight digital neuromorphic processor embedding a stochastic version of the spike-driven synaptic plasticity (S-SDSP) learning rule and a hierarchical routing fabric for large-scale chip interconnection. The MorphIC SNN processor embeds a total of 2k leaky integrate-and-fire (LIF) neurons and more than two million plastic synapses for an active silicon area of 2.86mm$^2$ in 65nm CMOS, achieving a high density of 738k synapses/mm$^2$. MorphIC demonstrates an order-of-magnitude improvement in the area-accuracy tradeoff on the MNIST classification task compared to previously-proposed SNNs, while having no penalty in the energy-accuracy tradeoff.
\end{abstract} 


\IEEEpeerreviewmaketitle

\section{Introduction} \label{sec_intro}

\begin{figure}[!t]
\centering
\noindent\includegraphics[width=0.95\columnwidth]{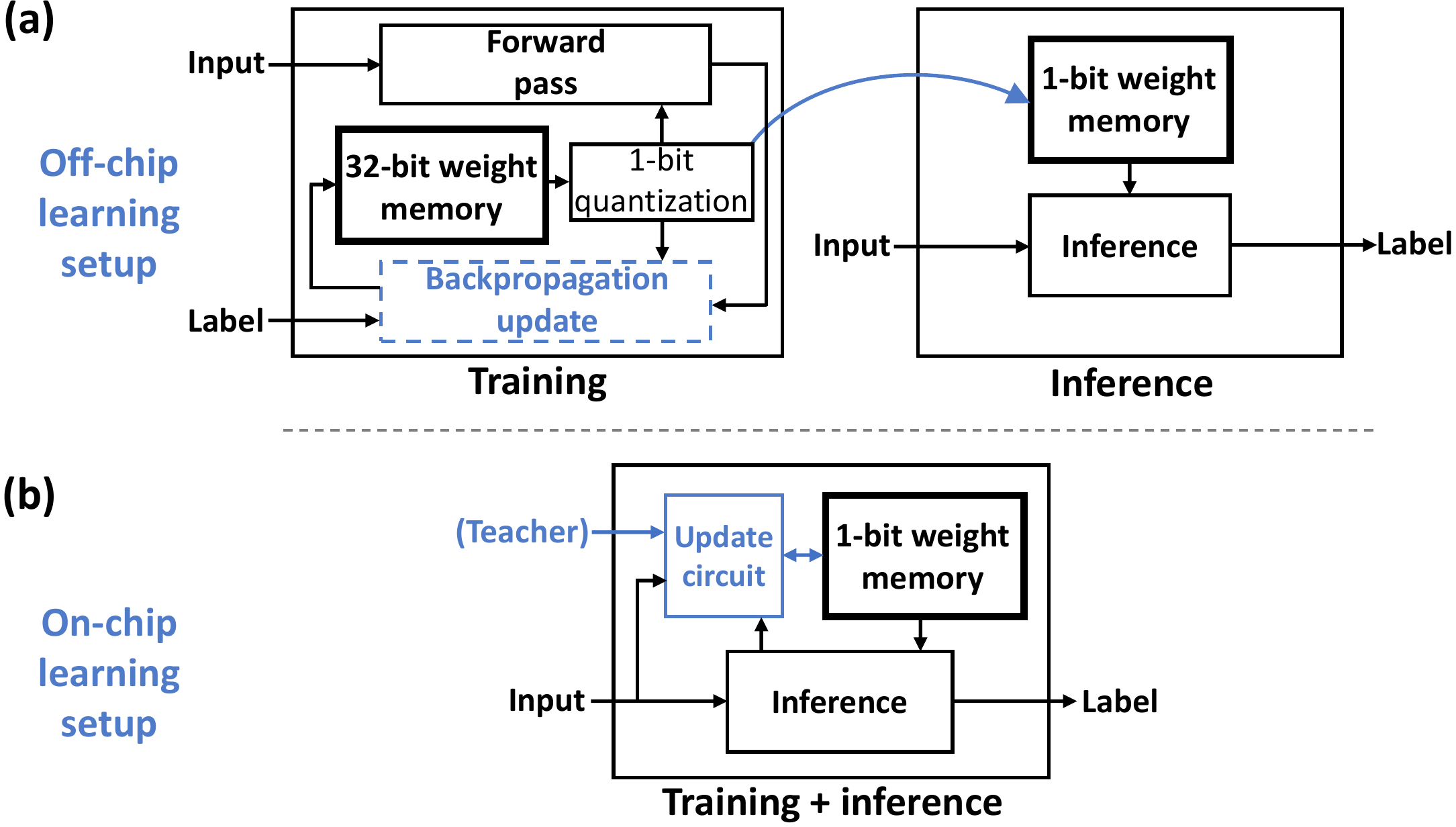}
\caption{Learning strategies for binary-weight neural networks. (a)~Quantization-aware off-chip learning setup: binary weights are used during the forward pass while full-resolution weights are kept for backpropagation updates~\cite{Courbariaux16}. Training is carried out in an off-chip high-performance optimizer, while inference is carried out in the power- and resource-constrained device. (b)~On-chip online learning setup, where data-driven weight updates are carried out in parallel with inference in the power- and resource-constrained device. A teacher signal is required for supervised online learning, whereas teacher-less learning is unsupervised.\vspace*{-1.5mm}}%
\label{fig_motivation}
\end{figure}

\IEEEPARstart{T}{he} massive deployment of neural network accelerators as inference devices is currently hindered by the memory footprint and power consumption required for high-accuracy classification~\cite{Whatmough17}. Two trends are being explored in order to solve this issue. The first trend consists in optimizing current artificial neural network (ANN) and convolutional neural network (CNN) architectures. Weight quantization down to binarization is a promising approach as it allows simplifying the operations and minimizing the memory footprint, thus avoiding the high energy cost of off-chip memory accesses if all the weights can be stored into on-chip memory~\cite{Moons17}. The accuracy drop induced by quantization can be mitigated to acceptable levels for many applications with the use of quantization-aware training techniques that propagate binary weights during the forward pass and keep full-resolution weights for backpropagation updates~\cite{Courbariaux16}. The associated off-chip learning setup for quantization-aware training is shown in Fig.~\ref{fig_motivation}(a): this strategy allows binary-weight neural networks to perform inference with a favorable energy-area-accuracy tradeoff, as recently demonstrated~\mbox{by binary CNN chips (e.g., \cite{Andri18,Moons18,Bankman18}).}

The second trend consists in changing the neural network architecture and data representation, which is currently being explored with bio-inspired spiking neural networks (SNNs) as a power-efficient neuromorphic processing alternative for sparse event-based data streams~\cite{Poon11}. Embedded online learning is a key feature in SNNs as it enables on-the-fly adaptation to the environment~\cite{Azghadi14}. Moreover, by avoiding the use of an off-chip optimizer, on-chip online learning allows SNNs to target applications that are power- and resource-constrained during both the training and the inference phases, as shown in Fig.~\ref{fig_motivation}(b). Spike-based online learning is an active research area, both in the development of new rules for high-accuracy learning in multi-layer networks (e.g.,~\cite{Zheng17,Mostafa17,Neftci17,Zenke18}) and in the demonstration of silicon implementations in applications such as unsupervised learning for image denoising and reconstruction~\cite{Knag15,Chen18}. However, these approaches currently rely on multi-bit weights.

These two trends mostly evolve in parallel as only three chips have been proposed previously to leverage the density and power advantage of binary weights with SNNs. First, the TrueNorth chip proposed by IBM is the largest-scale neuromorphic chip with 1M neurons and 256M 1-bit synapses, however it does not embed online learning~\cite{Akopyan15}. Second, the Loihi chip recently proposed by Intel has a configurable synaptic resolution that can be reduced to 1 bit and embeds a programmable co-processor for on-chip learning, though not demonstrated using a binary synaptic resolution to the best of our knowledge~\cite{Davies18}. Finally, Seo~\textit{et~al.} propose a stochastic version of the spike-timing-dependent plasticity (S-STDP) rule for online learning in binary synapses~\cite{Seo11}. However, S-STDP requires the design of a custom transpose SRAM memory with both row and column accesses, which severely degrades the density advantage of their approach.

It has been demonstrated in~\cite{Frenkel17} that the spike-dependent synaptic plasticity (SDSP) learning rule proposed by Brader~\textit{et~al.} in~\cite{Brader07} allows for a more efficient resource usage than STDP: all the information necessary for learning is available in the post-synaptic neuron at pre-synaptic spike time. SDSP requires neither an expensive local synaptic storage of spike timings nor a custom SRAM with both row and column accesses. Beyond a low implementation overhead, SDSP also embeds a stop-learning overfitting-prevention mechanism, whose efficiency is conditioned by a proper selection of the parameter values~\cite{Frenkel19}. In this work, we propose an efficient stochastic implementation of SDSP compatible with standard high-density foundry SRAMs in order to leverage embedded online learning in binary-weight SNNs.

Beyond plasticity, a second key aspect of spiking neural networks lies in connectivity. The brain organization in \textit{small-world networks} with dense local connectivity and sparse long-range wiring leads to efficient clustering of neuronal activity and hierarchical information encoding~\cite{Bassett06}. Network-on-chip (NoC) design applied to multi-core SNNs is thus an active research topic~\cite{Akopyan15,Davies18,Moradi18,Park17,Navaridas09,Benjamin14,Schemmel10}. In this work, we propose a hierarchical combination of mesh-based routing for inter-chip connectivity, star-based routing for intra-chip inter-core connectivity and crossbar-based routing for local intra-core connectivity. We store all the connectivity information locally in the neuron memory to enable memory-less routers that do not require local mapping table accesses. With only 27 connectivity bits per neuron, this low-memory hierarchical connectivity allows reaching biologically-realistic fan-in and fan-out values of 1k and 2k neurons, respectively.

We demonstrate this two-fold approach with MorphIC, a quad-core digital neuromorphic processor: stochastic SDSP (S-SDSP) is combined with a hierarchical routing fabric for large-scale plastic connectivity.  MorphIC was prototyped in 65nm CMOS and embeds 2k leaky integrate-and-fire (LIF) neurons and more than 2M synapses in an active silicon area of 2.86mm$^2$, therefore achieving a high density of 738k 1-bit online-learning synapses per mm$^2$. It results in an order-of-magnitude density improvement compared to the only previously-proposed binary-weight online-learning SNN processor from Seo~\textit{et~al.}~\cite{Seo11}. On the MNIST image recognition task~\cite{LeCun98}, MorphIC achieves a test set accuracy of 97.8\% with offline-trained binary weights. It demonstrates an order-of-magnitude improvement in the area-accuracy tradeoff compared to other SNNs, while having no penalty in the energy-accuracy tradeoff using rank order coding. Embedded online learning is validated by learning to discriminate eight patterns with S-SDSP. This paper extends~\cite{Frenkel19b} with detailed circuit, architectural and implementation aspects, while providing extended discussion of the measurement results compared to state-of-the-art neuromorphic chips.%

The remainder of this paper is structured as follows. The architecture and implementation of the MorphIC SNN processor are presented in Section~\ref{sec_arch}, together with detailed descriptions of the hierarchical event routing infrastructure and S-SDSP learning rule. The specifications, measurements and benchmarking results are provided in Section~\ref{sec_results}. Finally, the presented results are discussed in Section~\ref{sec_disc}.

\begin{figure}[!t]
\centering
\noindent\includegraphics[width=0.98\columnwidth]{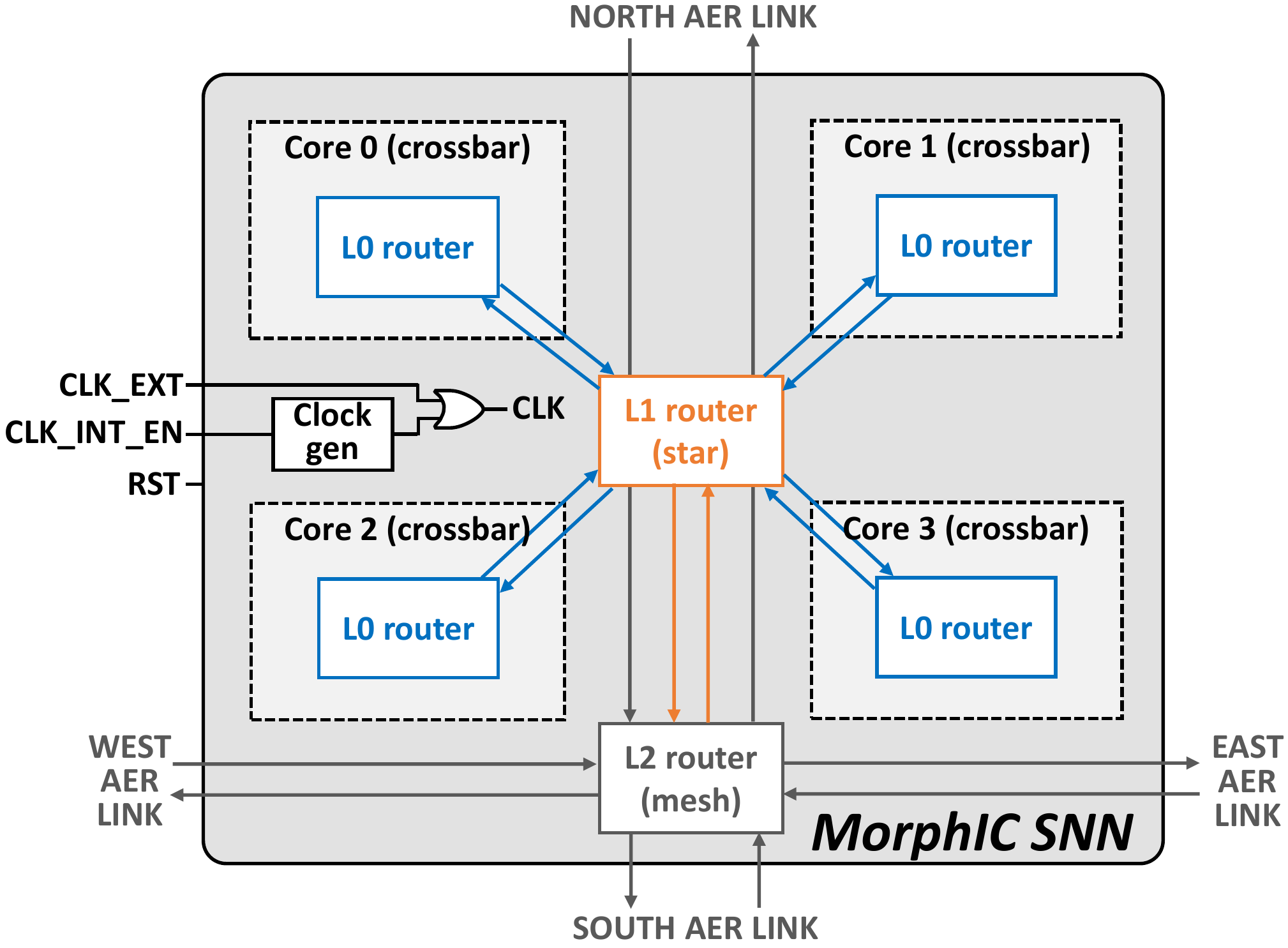}
\caption{Block diagram of the MorphIC quad-core neuromorphic processor.}%
\label{fig_morphic_snn}
\end{figure}

\begin{figure}[!t]
\centering
\noindent\includegraphics[width=0.90\columnwidth]{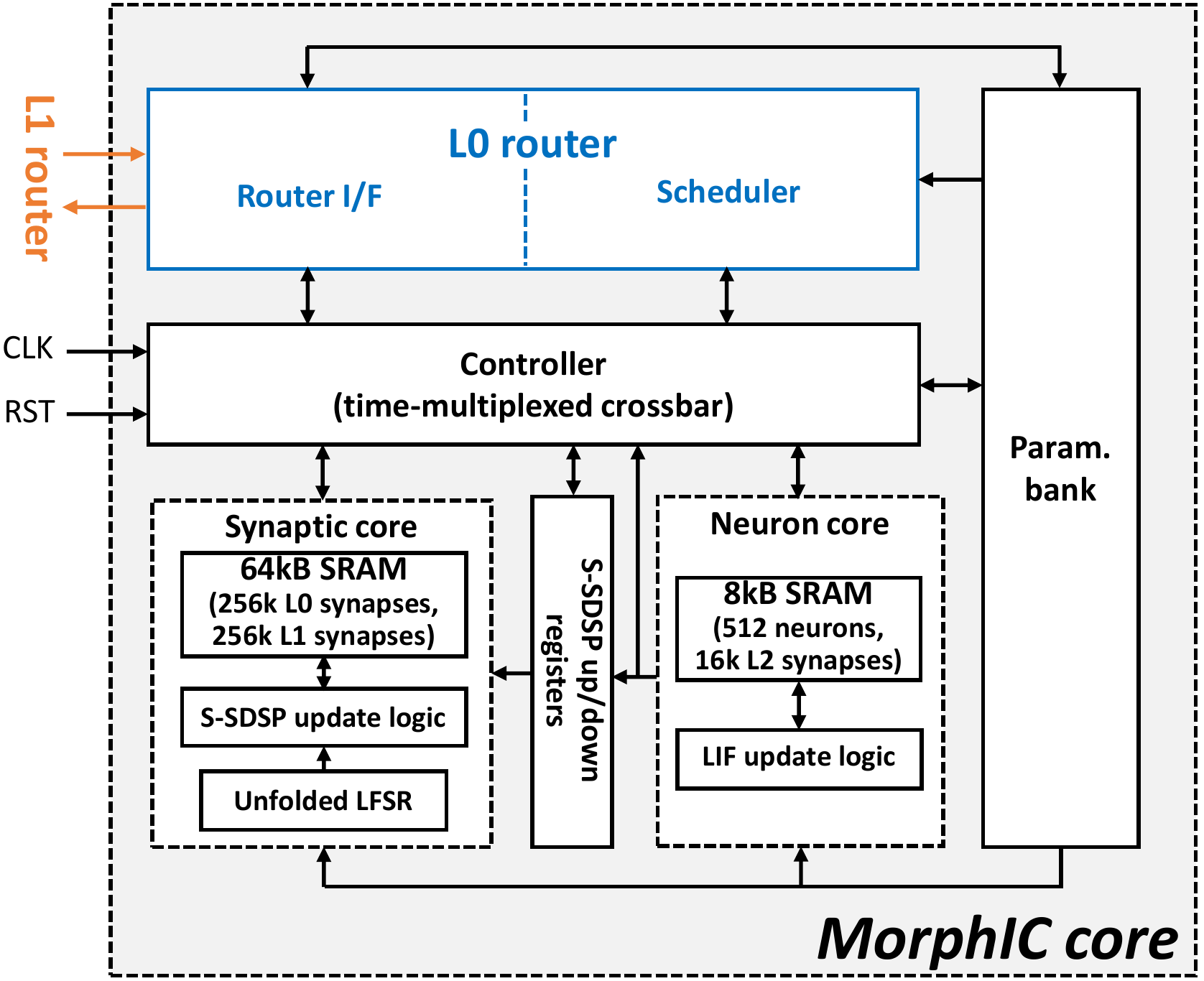}
\caption{Block diagram of a MorphIC core. Each core features 512 LIF neurons and 528k binary-weight synapses with embedded S-SDSP-based online learning.}%
\label{fig_morphic_core}
\end{figure}

\begin{figure*}[!t]
\centering
\setcounter{figure}{4}
\noindent\includegraphics[width=0.965\textwidth]{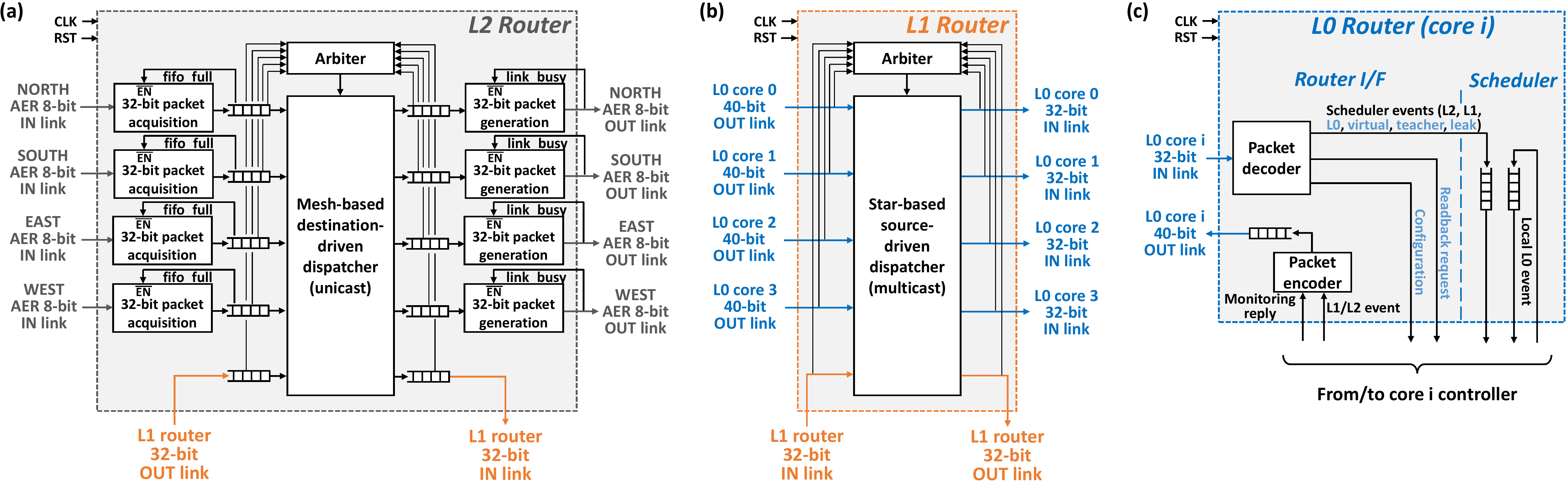}
\caption{Architecture of the hierarchical three-level event routing fabric of MorphIC. (a) The level-2 (L2) router handles high-level inter-chip connectivity with four bidirectional address-event-representation (AER) links, events are dispatched following a unicast mesh-based strategy. Packet buffering in FIFOs ensures that all links can operate independently. (b) The level-1 (L1) router handles mid-level intra-chip inter-core connectivity with four local links, one for each MorphIC core. Events are dispatched following a multicast star-based strategy. (c) The level-0 (L0) router handles low-level connectivity, it decodes incoming packets and sorts them toward either the controller or the scheduler of the local core. When a local neuron configured for L1 and/or L2 outward connectivity spikes, all its connectivity information is encapsulated in a routing packet before exiting the L0 router. Event types indicated in light blue are testbench-type events that cannot be generated by MorphIC chips.}
\label{fig_noc_arch}
\end{figure*}

\begin{figure}[!t]
\centering
\setcounter{figure}{3}
\noindent\includegraphics[width=0.98\columnwidth]{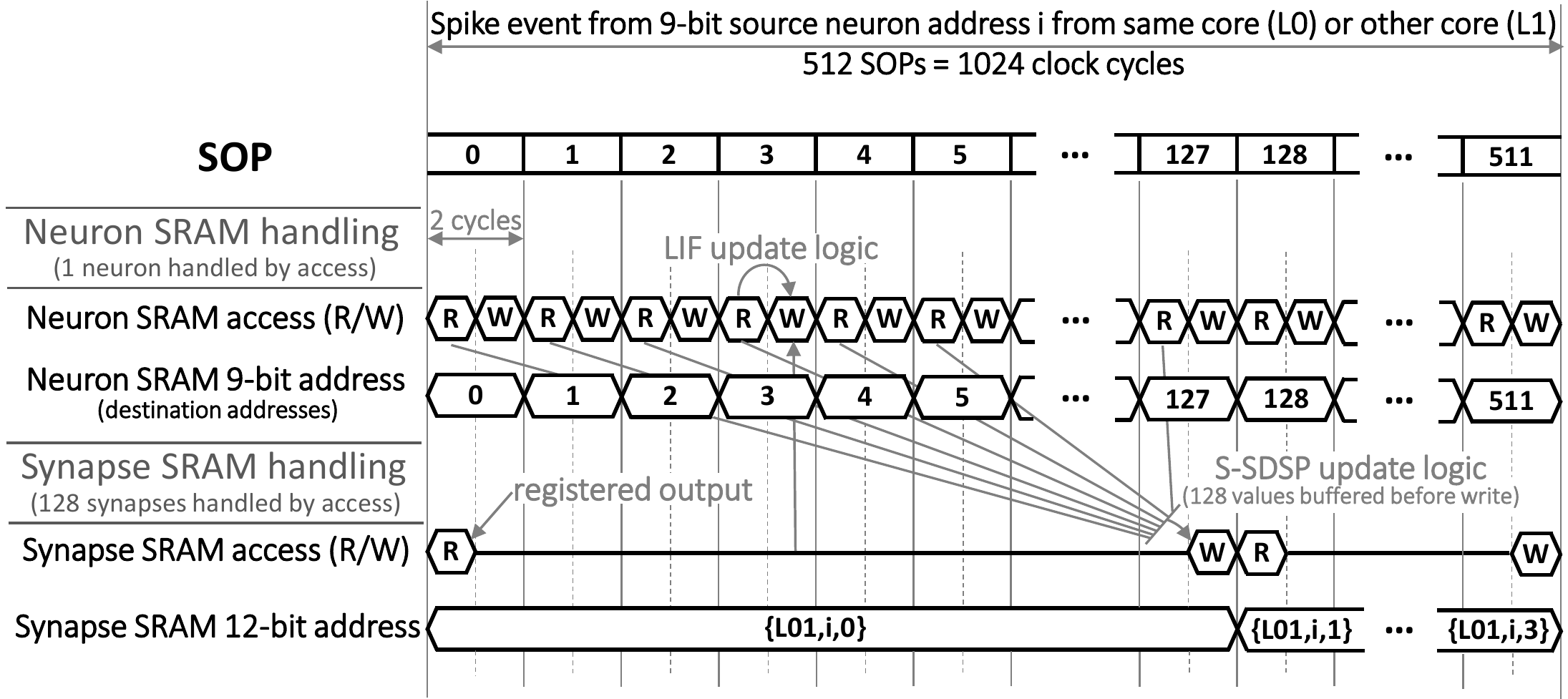}
\caption{Timing diagram of the crossbar operation in a MorphIC core, adapted from the time multiplexing scheme we previously proposed for the ODIN SNN in~\cite{Frenkel19}, illustrating the time-multiplexed crossbar operation for a spike event from 9-bit source neuron address $i$, leading to 512 synaptic operations (SOPs). Each SOP lasts two clock cycles. The core controller goes sequentially through all the local 512 neurons, it first reads their state in the local SRAM memory and then writes back the updated state retrieved from the leaky integrate-and-fire (LIF) update logic. The synapse SRAM has 128-bit words for density purposes: as MorphIC has 1-bit synapses, 128 synapses are handled by access and stochastic SDSP (S-SDSP) updates are buffered before being written back to the synapse SRAM memory. Depending on whether the source neuron was on the local core (L0 connectivity) or on another core from the same MorphIC chip (L1 connectivity), the MSB of the synapse SRAM address (\texttt{L01} flag bit) selects whether L0 or L1 synapses are accessed.}
\label{fig_clock_diagram}
\end{figure}

\section{Architecture and Implementation} \label{sec_arch}

A block diagram of the synchronous digital MorphIC quad-core spiking neuromorphic processor is shown in Fig.~\ref{fig_morphic_snn}, illustrating its \mbox{hierarchical} routing fabric for large-scale chip interconnection. Level-2~(L2) routers handle inter-chip connectivity, level-1~(L1) routers handle inter-core connectivity and level-0~(L0) routers handle intra-core connectivity~(Section~\ref{sec_hierout}). The clock can be either provided externally or generated internally using a configurable-length ring oscillator. A block diagram of the MorphIC core is shown in Fig.~\ref{fig_morphic_core}: each core embeds 512 leaky integrate-and-fire (LIF) neurons configured as a crossbar array with 256k L0 1-bit synapses and 256k L1 1-bit synapses, while 16k L2 synapses can be accessed independently. Each synapse embeds online learning with a stochastic implementation of the spike-dependent synaptic plasticity (S-SDSP) learning rule~(Section~\ref{sec_ssdsp}). Each axon can be configured to multiply its associated synaptic weights by a factor of 1, 2, 4 or 8. Time multiplexing is used to increase the neuron and synapse densities by using shared update circuits and storing neuron and synapse states to local SRAM memory, based on the strategy we previously proposed for the ODIN SNN in~\cite{Frenkel19}. Fig.~\ref{fig_clock_diagram} illustrates the time-multiplexed crossbar operation of a MorphIC core when it processes a spike event from a neuron in the local core (L0 connectivity) or from a neuron in another core in the same chip (L1 connectivity). The core controller goes sequentially through all the 512 local neurons, leading to 512 synaptic operations (SOPs), and handles the local SRAM memory accesses accordingly. As L2 events target a specific synapse of a neuron (Section~\ref{sec_hierout}), they lead to a single SOP.

\subsection{Hierarchical event routing} \label{sec_hierout}

\begin{figure}[!t]
\centering
\vspace*{-1.7mm}
\setcounter{figure}{5}
\noindent\includegraphics[width=0.96\columnwidth]{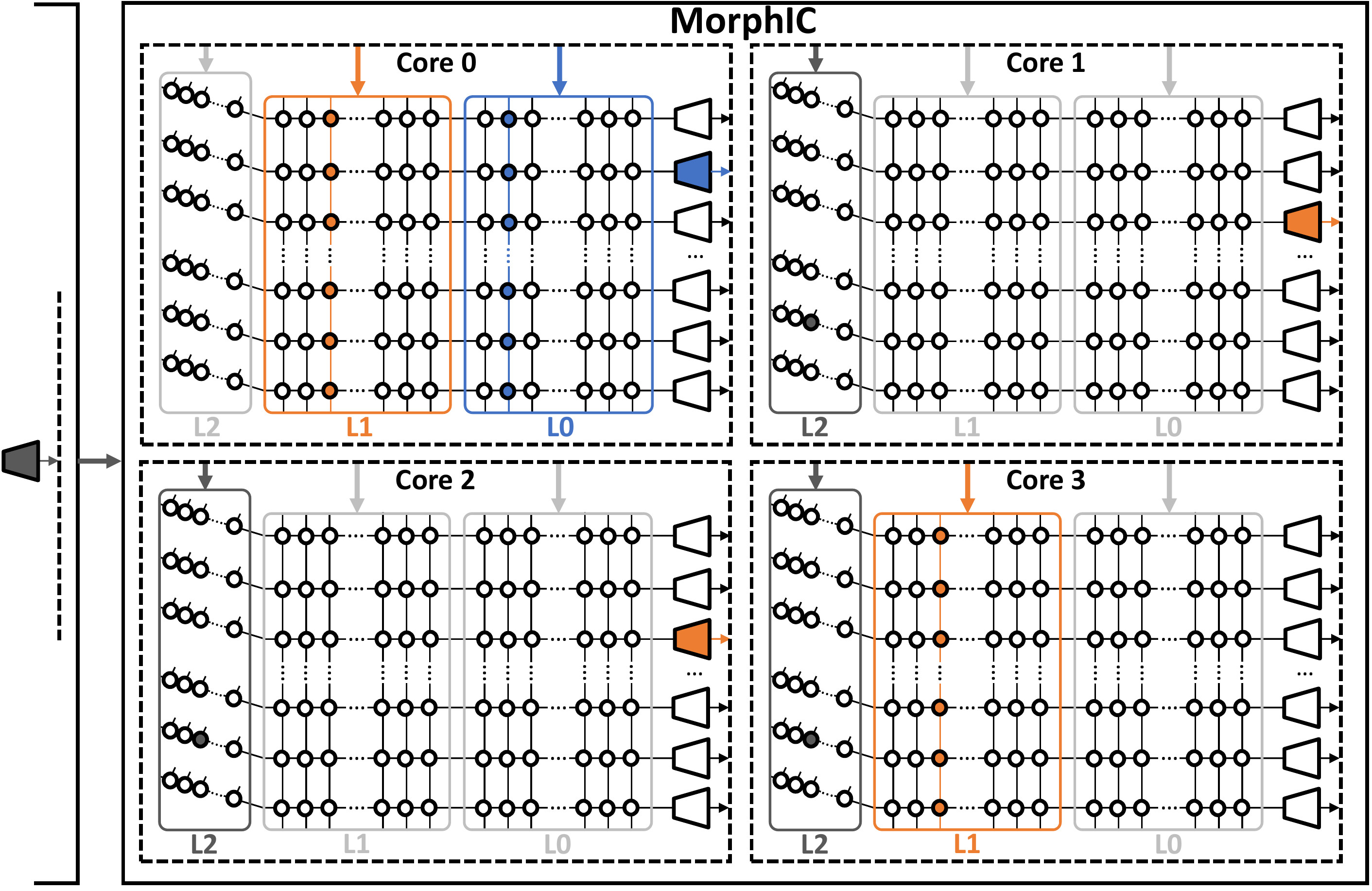}
\caption{Examples of L0, L1 and L2 connectivity handling at the core level. Blue: L0 connectivity inside core 0, following a typical crossbar operation. Orange: L1 connectivity from neurons in cores 1 and 2 to cores 0 and 3, following a crossbar operation in the destination cores. In this example, as the source neurons have identical 9-bit addresses, they map to the same L1 synapses in the destination cores. Gray: L2 connectivity from a neuron in another MorphIC chip to a specific L2 synapse of a target neuron, broadcasted to cores 1, 2 and 3 of the destination chip.\vspace*{-2mm}}
\label{fig_noc_illus}
\end{figure}

Clustering groups of neurons with dense local and sparse long-range connectivity allows minimizing memory requirements while keeping flexibility and scalability~\cite{Moradi18}. This organization is found in the brain and is known as \textit{small-world networks}. Hierarchy is therefore a key concept in SNN event routing infrastructures for large-scale networks~\cite{Akopyan15,Davies18,Moradi18,Park17,Navaridas09,Benjamin14,Schemmel10}. MorphIC uses a heterogeneous hierarchical routing fabric with different router types at each level, as shown in Fig.~\ref{fig_noc_arch}: the L2 router follows a unicast mesh-based dimension-ordered destination-driven operation (Section~\ref{sssec_l2}), the L1 router follows a multicast star-based source-driven operation (Section~\ref{sssec_l1}) while the L0 router handles decoding and encoding of the different packet types for local core crossbar-based processing (Section~\ref{sssec_l0}). The individual routing levels and their combination do not contain cyclic path dependencies and are thus deadlock-free. This heterogeneous event routing infrastructure allows for the three connectivity patterns illustrated in Fig.~\ref{fig_noc_illus}, depending on the source neuron location:
\begin{itemize}
\item The source neuron targets neurons in the same core (L0 connectivity): the time-multiplexed crossbar approach of Fig.~\ref{fig_clock_diagram} is followed with the local L0 synapses (e.g., blue pattern in core 0 in Fig.~\ref{fig_noc_illus}).
\item The source neuron targets neurons in any combination of other cores in the same chip (L1 connectivity): the time-multiplexed crossbar approach of Fig.~\ref{fig_clock_diagram} is followed with the L1 synapses of the destination cores. The same L1 synapses are shared with up to three cores (e.g., orange pattern from source neurons in cores 1 and 2 to destination cores 0 and 3 in Fig.~\ref{fig_noc_illus}).
\item The source neuron is located in another MorphIC chip (L2 connectivity): the target is a specific L2 synapse address in any combination of cores in one destination chip (e.g., gray pattern from a source neuron retrieved from the \textit{West} link toward identical L2 synapse addresses in cores 1, 2 and 3 in Fig.~\ref{fig_noc_illus}). As each neuron has 32 L2 synapses, an L2 synapse address has a width of 14 bits (9 bits for the neuron, 5 bits for the L2 synapse).
\end{itemize}

Each neuron of MorphIC can use any combination of the aforementioned three types of L0, L1 and L2 connectivities, which allows reaching a fan-in of 512 (L0) + 512 (L1) + 32 (L2) and a fan-out of 512 (L0) + 3$\times$512 (L1) + 4 (L2).

\begin{figure}[!t]
\centering
\vspace*{-1mm}
\noindent\includegraphics[width=0.99\columnwidth]{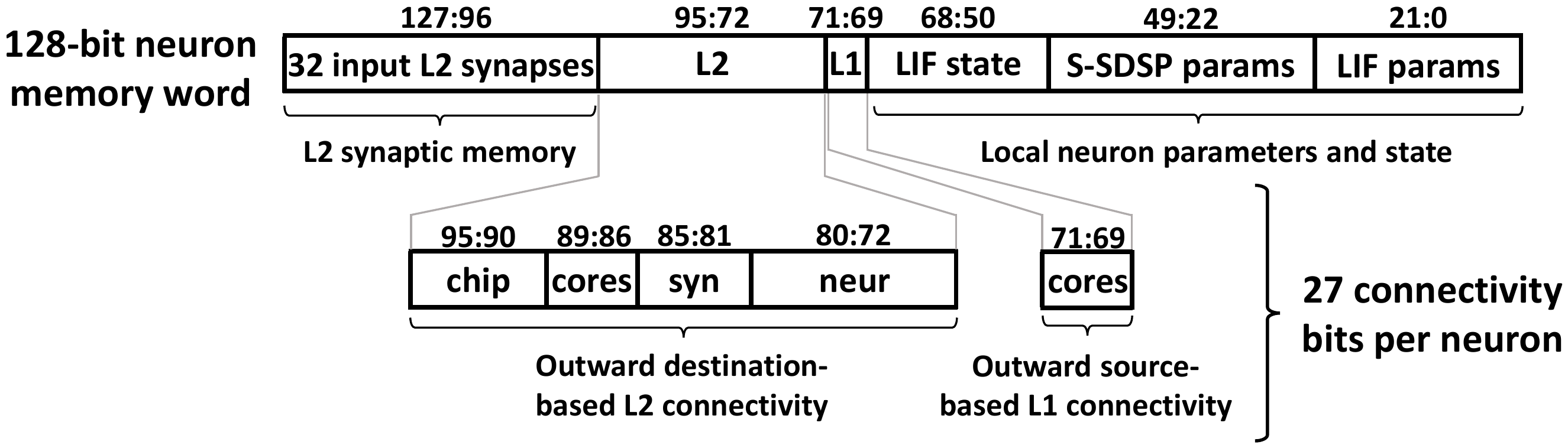}
\caption{Neuron memory map: structure of a 128-bit word in the neuron SRAM memory. Each word contains the parameters, state, outward connectivity and the 32 1-bit input L2 synapses of a neuron. At runtime, only the LIF state (i.e.~an 11-bit membrane potential, a 4-bit Calcium variable and a 4-bit counter to emulate Calcium leakage) and the L2 synapses can be modified by the LIF and S-SDSP update logic blocks~(Fig.~\ref{fig_morphic_core}), respectively. The L2 and L1 connectivity fields occupy a total of only 27 bits per neuron.}%
\label{fig_memory_map}
\end{figure}

 The entire connectivity of a network of MorphIC chips is determined by only 27 connectivity bits per neuron, which are stored in the neuron 8-kB SRAM memories located inside each core (Fig.~\ref{fig_morphic_core}). It consists of 512 128-bit words, one word for each of the 512 LIF neurons per core, whose structure is outlined in Fig.~\ref{fig_memory_map}. Destination-based L2 connectivity requires 24 bits in total: the 6-bit \texttt{chip} field stores 3-bit $dx$ and $dy$ fields encoding the destination chip (Section~\ref{sssec_l2}), the 4-bit \texttt{cores} field encodes the combination of target cores and the 5-bit \texttt{syn} and the 9-bit \texttt{neur} fields encode the 14-bit L2 synapse address. Source-based L1 connectivity requires only 3 bits per neuron in order to target any combination of the other cores in a MorphIC chip. Except if disabled in the core parameter bank, L0 crossbar connectivity is automatic and does not require further connectivity information. As all the connectivity information is decentralized next to the neurons and then encapsulated in the event packets, the routers do not require local or external mapping tables: they are memory-less beyond simple packet buffering. Let us now discuss the architectural details of the L2, L1 and L0 routers.%

\subsubsection{Level-2 (L2) router} \label{sssec_l2} The L2 router (Fig.~\ref{fig_noc_arch}(a)) handles high-level inter-chip connectivity with four links along the \textit{North}, \textit{South}, \textit{East} and \textit{West} directions that operate independently and in parallel. Events from/to the four chip-level links and from/to the L1 router are buffered into FIFOs before being dispatched following a standard unicast mesh-based strategy with dimension-ordered routing (i.e. $x$ direction before $y$ direction). Two $dx$ and $dy$ fields in the chip-level packet contain the information necessary for destination-based routing. $dx$ and $dy$ have a 3-bit width each (one sign bit, two data bits), which allows routing packets to up to three MorphIC chips in any direction. At each \textit{East} or \textit{West} (resp. \textit{North} or \textit{South}) hop, the L2 router decrements the value of the $dx$ (resp. $dy$) data field. When both $dx$ and $dy$ are zero, the packet is then forwarded to the L1 router. Distance information $d$ is also maintained separately in the event packet: $d$ is 0 for local L0 events and 1 for events received from local L1 connectivity, it then increases for each L2 hop up to a maximum of 7 for events received from a chip located at $dx$=$\pm$3 and $dy$=$\pm$3. As synapses at all routing levels of MorphIC embed online learning~(Section~\ref{sec_ssdsp}), the probability of synaptic weight update can be modulated by the distance information, following a small-world network modeling strategy. To the best of our knowledge, this is the first SNN to propose online hierarchical learning.

The mesh-based dispatcher is controlled by an arbiter, which can be configured either for round-robin or for priority-based operation. Round-robin operation, by cycling through each link independently of the FIFO usage, guarantees a maximum latency for packet processing, while priority-based operation is a greedy approach that allocates processing time to the most active links based on the current FIFO usage.

\begin{figure}[!t]
\centering
\noindent\includegraphics[width=0.9\columnwidth]{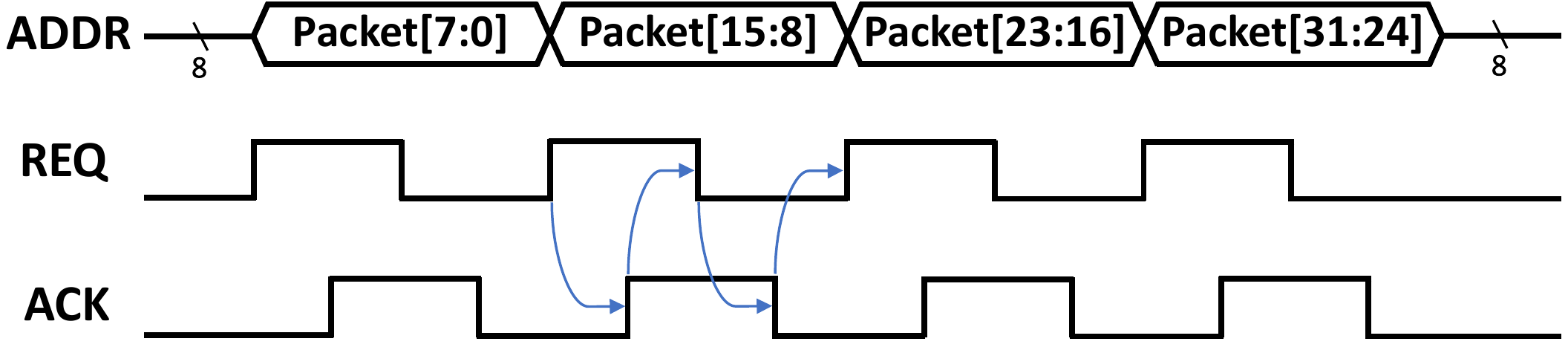}
\caption{32-bit packet transmission multiplexed into four 8-bit AER transactions at the L2 level. As double-latching synchronization barriers are used on the receiver \texttt{REQ} line, the \texttt{ADDR} data can be asserted at the same time as the \texttt{REQ} line on the sender side.}
\label{fig_aer_l2}
\end{figure}

Links in each direction consist of two address-event representation (AER) buses, a sender and a receiver, for a total of eight AER buses per MorphIC chip. AER is a \textit{de facto} standard for spiking neural network connectivity as it allows high-speed asynchronous communication of spike events between chips using a four-phase handshake protocol~\cite{Mortara94, Boahen00}. The MorphIC design being pad-limited, the width of the AER buses has been reduced to 8 bits. Transmission and reception of 32-bit event packets are thus multiplexed into four 8-bit AER transactions, as illustrated in Fig.~\ref{fig_aer_l2}.  In order to ensure an asynchronous operation of the AER buses between MorphIC chips, double-latching synchronization barriers have been placed on the receiver \texttt{REQ} and sender \texttt{ACK} handshake lines to limit metastability issues. As the pads are the speed bottleneck for off-chip L2 packet routing (Table~\ref{table_specs}), L2 packet activity should be sparse compared to L1 and L0 activity: L2 events should thus represent high-level features, as illustrated in the experiments outlined in Section~\ref{sec_results}. The L2 routing speed could be improved by using a 2-phase handshake AER variant instead of the standard 4-phase handshake.

\subsubsection{Level-1 (L1) router} \label{sssec_l1} The L1 router (Fig. \ref{fig_noc_arch}(b)) handles mid-level intra-chip inter-core connectivity with the four local MorphIC cores. This router is based on a star topology and relies on a simple dispatcher that multicasts events to local cores following a source-based approach. It does not contain any FIFO buffering as awaiting packets are already buffered in the L2 and L0 routers. An arbiter controls the dispatcher following a configurable round-robin or greedy priority-based operation, similarly to the L2 router.

The L1 router is at the center of the hierarchy. For neuron events from local cores (i.e. ascending-hierarchy events), it handles multicasting to any combination of the other cores toward L1 synapses and/or forwarding to the L2 router toward another MorphIC chip. For events retrieved from the L2 router (i.e. descending-hierarchy events), it handles multicasting to any combination of the MorphIC cores toward L2 synapses.

\subsubsection{Level-0 (L0) router} \label{sssec_l0} The L0 router (Fig. \ref{fig_noc_arch}(c)) handles low-level intra-core connectivity. This router is divided into two blocks: an interface and a scheduler. The interface handles packet decoding and encoding from/to the L1 router. The packet decoder segments input packets into different types:
\begin{itemize}
\item \textit{configuration} packets are used to program the local neuron and synapse SRAMs and the core parameter bank (Fig.~\ref{fig_morphic_core}), they are handled by the controller,
\item \textit{monitoring request} packets query one byte from the neuron or synapse SRAM, they are handled by the controller,
\item \textit{scheduler} events are buffered by a FIFO in the core scheduler, they include \textit{L2} events targeting a single L2 synapse, \textit{L1} events targeting L1 synapses, \textit{L0} events targeting L0 synapses, \textit{virtual} events that directly update a neuron without accessing any physical synapse, \textit{teacher} events that control the S-SDSP supervision mechanism through the neuron Calcium variables (Section~\ref{sec_ssdsp}) and \textit{leak} events that drive the LIF leakage time constant.
\end{itemize}

Locally-generated \textit{L0} events are buffered directly in a scheduler FIFO, they are not visible from the L1/L2 router hierarchy. Locally-generated events that need to go up the router hierarchy are handled by the packet encoder:
\begin{itemize}
\item \textit{monitoring reply} packets contain the neuron or the synapse state byte previously queried by a \textit{monitoring request} packet,
\item \textit{L1/L2} events forward the L1 and L2 connectivity information of a source neuron to the L1 router.
\end{itemize}

\begin{figure*}[!t]
\centering
\vspace*{-2.5mm}
\setcounter{figure}{9}
\noindent\includegraphics[width=0.66\textwidth]{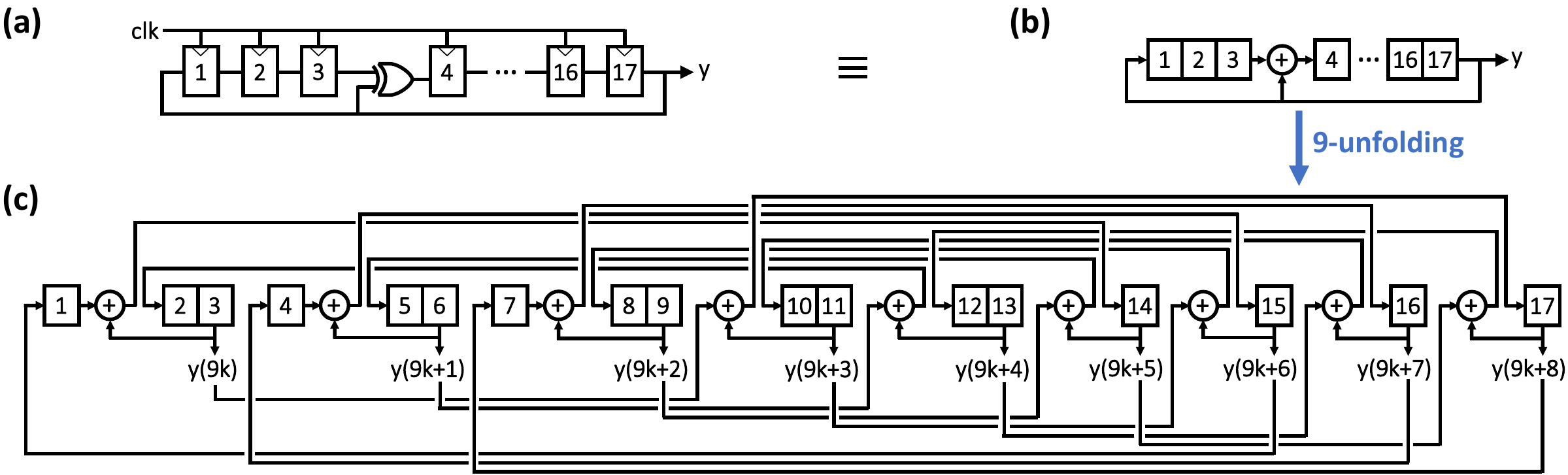}
\caption{(a) Circuit diagram of a Galois 17-bit LFSR with characteristic polynomial $x^{17}+x^3+1$. (b) Equivalent compact representation. (c) 9-unfolded 17-bit Galois LFSR for single-cycle 9-bit pseudo-random word generation for the S-SDSP online learning rule.\vspace*{-2mm}}%
\label{fig_ulfsr}
\end{figure*}

\subsection{Stochastic spike-dependent synaptic plasticity (S-SDSP)} \label{sec_ssdsp}

As the spike-timing-dependent plasticity (STDP) learning rule relies on the relative timing between pre- and post-synaptic spikes, it requires a local synaptic buffering of spike timings, which leads to critical overheads as buffering circuitry has to be replicated inside each synapse~\cite{Frenkel17}. In order to avoid this problem, the stochastic binary approach proposed by Seo~\textit{et~al.} in~\cite{Seo11} involves the design of a custom transpose SRAM with both row and column accesses to carry out STDP updates each time pre- and post-synaptic spikes occur. However, beyond increasing the design time, custom SRAMs do not benefit from DRC pushed rules for foundry bitcells and induce a strong area penalty compared to single-port high-density foundry SRAMs~\cite{Frenkel17}. Therefore, STDP cannot be implemented efficiently in silicon.

The spike-dependent synaptic plasticity (SDSP) learning rule~\cite{Brader07} avoids this drawback: the synaptic weight $w$ is updated each time a pre-synaptic event occurs, according to Eq.~(\ref{eq_sdsp}). The update depends solely on the state of the post-synaptic neuron at the time of the pre-synaptic spike, i.e. the membrane potential $V_\text{mem}$ compared to threshold $\theta_m$ and the Calcium concentration $\text{Ca}$ compared to thresholds $\theta_1$, $\theta_2$ and $\theta_3$. The Calcium concentration represents an image of the recent firing activity of the neuron, it disables SDSP updates for high and low post-synaptic neuron activities and helps prevent overfitting~\cite{Brader07}. A single-port high-density foundry SRAM can therefore be used for high-density time-multiplexed implementations. However, as SDSP relies on discrete positive and negative steps, it cannot be \mbox{applied directly to binary weights.}
\begin{equation}
\hspace*{-5mm}\begin{aligned}
\scalebox{0.95}[1]{
$\begin{cases}
w \rightarrow w + 1  &\text{if~~} V_{\text{mem}}(t_{\text{pre}})\geq\theta_{m}, ~\theta_1\leq \text{Ca}(t_{\text{pre}})<\theta_3 \\
w \rightarrow w - 1  &\text{if~~} V_{\text{mem}}(t_{\text{pre}})<\theta_{m}, ~\theta_1\leq \text{Ca}(t_{\text{pre}})<\theta_2
\end{cases}$}
\end{aligned}
\label{eq_sdsp}
\raisetag{18pt}
\end{equation}

Senn and Fusi proposed a bio-inspired stochastic learning rule for binary synapses in~\cite{Senn05}, where the update conditions rely on the total synaptic input of the post-synaptic neuron at the time of the pre-synaptic spike. However, this information is not easily available in time-multiplexed implementations: as shown in Fig.~\ref{fig_clock_diagram}, the \textit{destination} neurons are processed sequentially, while obtaining the total post-synaptic input of a neuron would require sequential processing of the \textit{source} neurons instead, which is incompatible with an event-driven operation. Therefore, we propose a stochastic spike-dependent synaptic plasticity (S-SDSP) learning rule suitable for binary weights, as formulated in Eq.~(\ref{eq_ssdsp}). It results from the fusion of the stochastic mechanism proposed in~\cite{Senn05} with the SDSP update conditions. $\zeta^+$ and $\zeta^\text{--}$ are binary random variables with probabilities $q^+$ and $q^\text{--}$ of being at 1, respectively. The synaptic weight $w_b$ therefore goes from 0 to 1 (resp. from 1 to 0) with probability $q^+$ (resp. $q^\text{--}$), depending on the update conditions. The Calcium concentration is implemented as a 4-bit variable, it is stored next to all S-SDSP parameters in the neuron SRAM~(Fig.~\ref{fig_memory_map}).
\begin{equation}
\hspace*{-5mm}\begin{aligned}
\scalebox{0.85}[1]{
$\begin{cases}
w_b \rightarrow w_b + \zeta^+(1-w_b)  &\text{if~~} V_{\text{mem}}(t_{\text{pre}})\geq\theta_{m}, ~\theta_1\leq \text{Ca}(t_{\text{pre}})<\theta_3 \\
w_b \rightarrow w_b - \zeta^\text{--}w_b  &\text{if~~} V_{\text{mem}}(t_{\text{pre}})<\theta_{m}, ~\theta_1\leq \text{Ca}(t_{\text{pre}})<\theta_2
\end{cases}$}
\end{aligned}
\label{eq_ssdsp}
\raisetag{18pt}
\end{equation}

\begin{figure}[!t]
\centering
\setcounter{figure}{8}
\noindent\includegraphics[width=0.87\columnwidth]{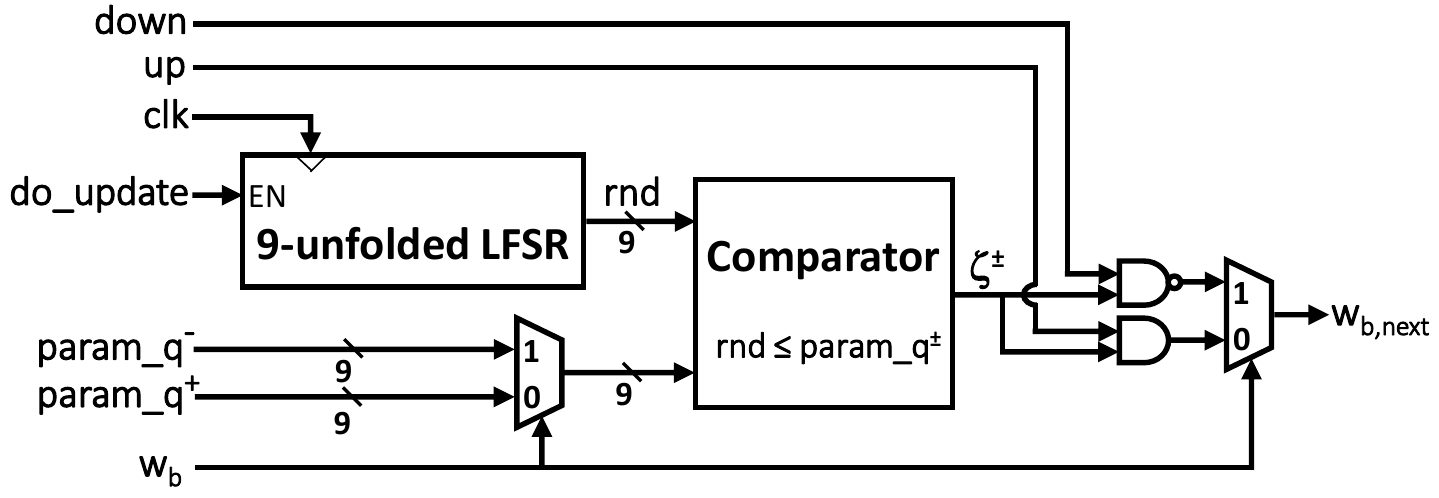}
\caption{Time-multiplexed S-SDSP update logic. \texttt{up} and \texttt{down} signals represent the values of the S-SDSP update conditions in Eq.~(\ref{eq_ssdsp}).\vspace*{-3mm}}%
\label{fig_ssdsp}
\end{figure}

The proposed S-SDSP update logic is shown in Fig.~\ref{fig_ssdsp}. The binary random variables $\zeta^\pm$ can be generated with $q^\pm$ probabilities using linear feedback shift register (LFSR)-based pseudo-random number generation. In order to generate $q^\pm$ with a resolution similar to the probabilities down to 0.01 used in~\cite{Senn05}, approximately 6 bits of resolution are required. Distance-based modulation of $q^\pm$ from small-world network modeling requires another 3 bits of resolution as the distance information ranges from 0 to 7~(Section~\ref{sec_hierout}). Therefore, we selected a 9-bit resolution for $q^\pm$ probabilities. As \mbox{S-SDSP} updates must be computed in a single clock cycle, it is possible to parallelize successive iterations of an LFSR by using the unfolding algorithm from~\cite{Parhi99}, as suggested in~\cite{Cheng06} to avoid instantiating parallel LFSRs and save switching power. The number of parallelized successive iterations is governed by the unfolding factor, which is 9 in this case. The unfolding process and the resulting unfolded LFSR are illustrated in Fig.~\ref{fig_ulfsr}. Unfolding leads the combinational logic resources (here, a single XOR gate) to be multiplied by the unfolding factor, while the LFSR period is divided by the unfolding factor. In order to avoid inducing correlation between synapses, the period of the unfolded LFSR must be one order of magnitude higher than the number of synapses per neuron. We thus selected a 17-bit depth for the LFSR to be unfolded (Fig.~\ref{fig_ulfsr}(a-b)). The 9-unfolded LFSR is shown in Fig.~\ref{fig_ulfsr}(c). The overhead incurred by the resulting S-SDSP update logic is negligible as it is shared with time multiplexing for all the L0, L1 and L2 synapses in a MorphIC core.

\begin{figure}[!t]
\centering
\setcounter{figure}{10}
\noindent\includegraphics[width=0.625\columnwidth]{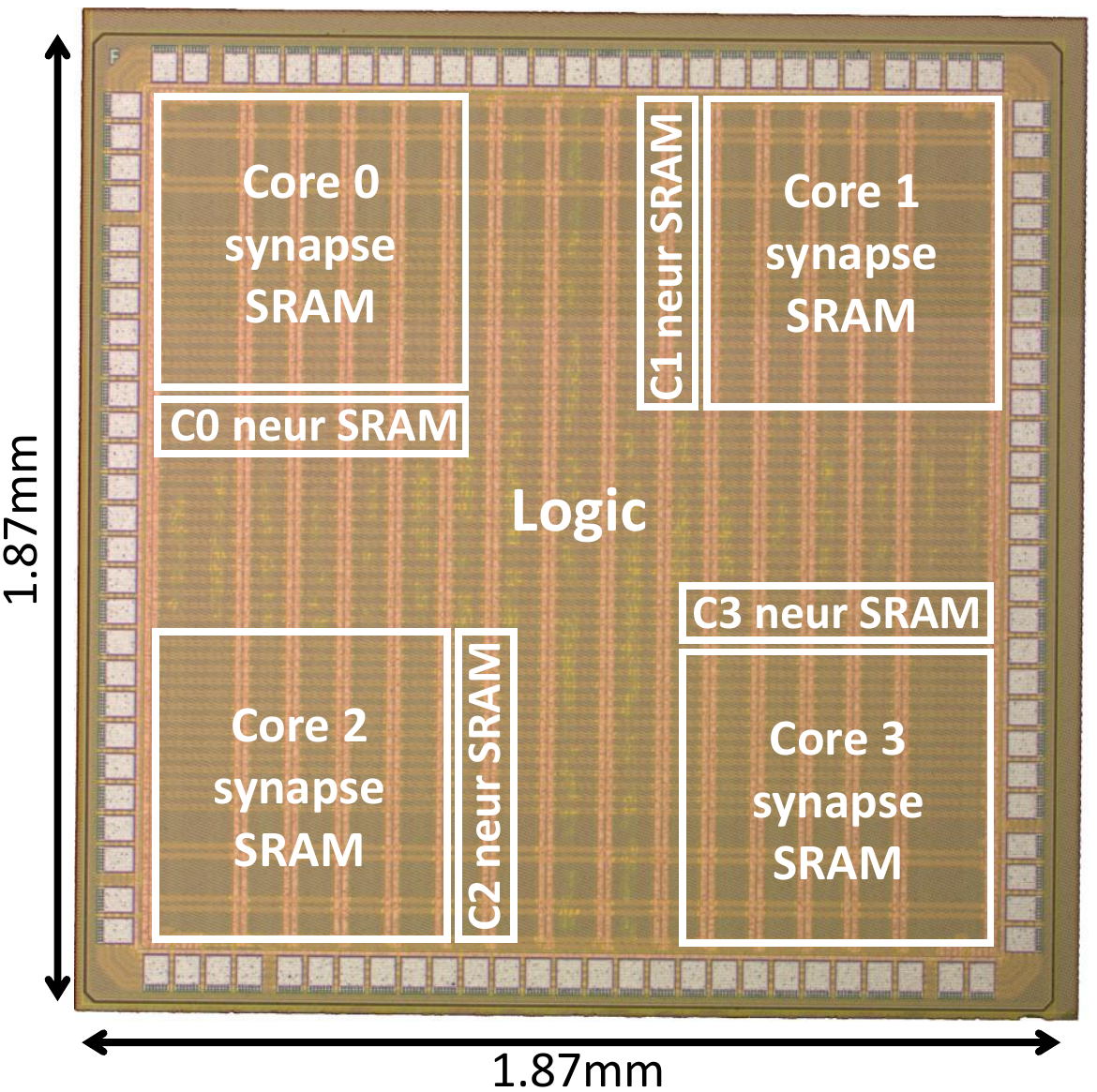}
\caption{MorphIC chip microphotograph, illustrating the floorplan of the neuron and synapse SRAM macros of each core.}
\label{fig_chip}
\end{figure}

\begin{table}[!t]
\caption{Specifications and measurements of MorphIC.}
\label{table_specs}
\renewcommand{\arraystretch}{0.9}
\centering
\resizebox{0.9\columnwidth}{!}{
\begin{tabular}{lcc}
\toprule%
Technology & \multicolumn{2}{c}{65nm LP CMOS} \\
Implementation & \multicolumn{2}{c}{Digital} \\
\multirow{3}{*}{Area} & \multicolumn{2}{c}{3.50mm$^2$ (chip, incl. pads)} \\
& \multicolumn{2}{c}{2.86mm$^2$ (chip, excl. pads)} \\
& \multicolumn{2}{c}{1.59mm$^2$ (SRAM macros)} \\
Total SRAM memory (type) & \multicolumn{2}{c}{256kB (syn), 32kB (neur)} \\
Number of cores & \multicolumn{2}{c}{4} \\
Total \# neurons (type) & \multicolumn{2}{c}{2048 (LIF)} \\
Total \# synapses (hier.) & \multicolumn{2}{c}{1M (L0), 1M (L1), 64k (L2)} \\
Fan-in (hier.) & \multicolumn{2}{c}{512 (L0), 512 (L1), 32 (L2)} \\
Fan-out (hier.) & \multicolumn{2}{c}{512 (L0), 3$\times$512 (L1), 4 (L2)} \\
Online learning & \multicolumn{2}{c}{Stochastic SDSP, 1-bit weights} \\
Time constant & \multicolumn{2}{c}{Biological to accelerated} \\
Supply voltage & 0.8V & 1.2V \\
Max. clock frequency & 55MHz & 210MHz \\
Max. acceleration$^\circ$ & 10$\times$ & 40$\times$ \\
Leakage power ($P_\text{leak}$) & 45$\mu$W & 190$\mu$W \\
Idle power ($P_\text{idle}$) & 41.3$\mu$W/MHz & 94.0$\mu$W/MHz \\
Energy per SOP ($E_\text{SOP}$) & 30pJ & 65pJ \\
Energy per L2 hop$^\ddagger$ & 9.0pJ & 20.3pJ \\
Energy per L1 hop$^*$ & 1.7pJ & 3.8pJ \\
L2 router bandwidth (AER) & 2.3Mpackets/s/link & 5.7Mpackets/s/link$^\dagger$ \\
L1 router bandwidth & 55Mpackets/s & 210Mpackets/s \\
Core bandwidth (max. $r_\text{SOP}$) & 27.5MSOP/s/core & 105MSOP/s/core \\
\bottomrule%
\end{tabular}}
\vspace*{0.5mm}

\scriptsize

$^\circ$ Compared to biological-time processing with all neurons assumed to spike at 10Hz.
$^\ddagger$ Excluding IO power.~~ $^*$ Simulation results.~~ $^\dagger$ Limited by pad delay at high speed.
\end{table}

\begin{table}[!t]
\caption{Silicon area breakdown (logic and SRAM without I/O pads).}
\label{table_area}
\renewcommand{\arraystretch}{0.85}
\centering
\resizebox{0.55\columnwidth}{!}{
\begin{tabular}{lll}
\toprule%
\multirow{4}{*}{Cores}& Synapses & 66.27\% \\
& Neurons & 16.61\% \\
& Parameters & 5.88\% \\
& Controller & 0.36\% \\
\hdashline
\rule{0pt}{2.3ex}   
\multirow{3}{*}{Routers} & L2 & 5.47\% \\
& L1 & 0.18\% \\
& L0 & 5.19\% \\
\hdashline
\rule{0pt}{2.3ex} 
Others & Clock generator & 0.04\% \\
\bottomrule%
\end{tabular}}%
\end{table}

\begin{figure*}[!t]
\centering
\vspace*{-1mm}
\noindent\includegraphics[width=0.94\textwidth]{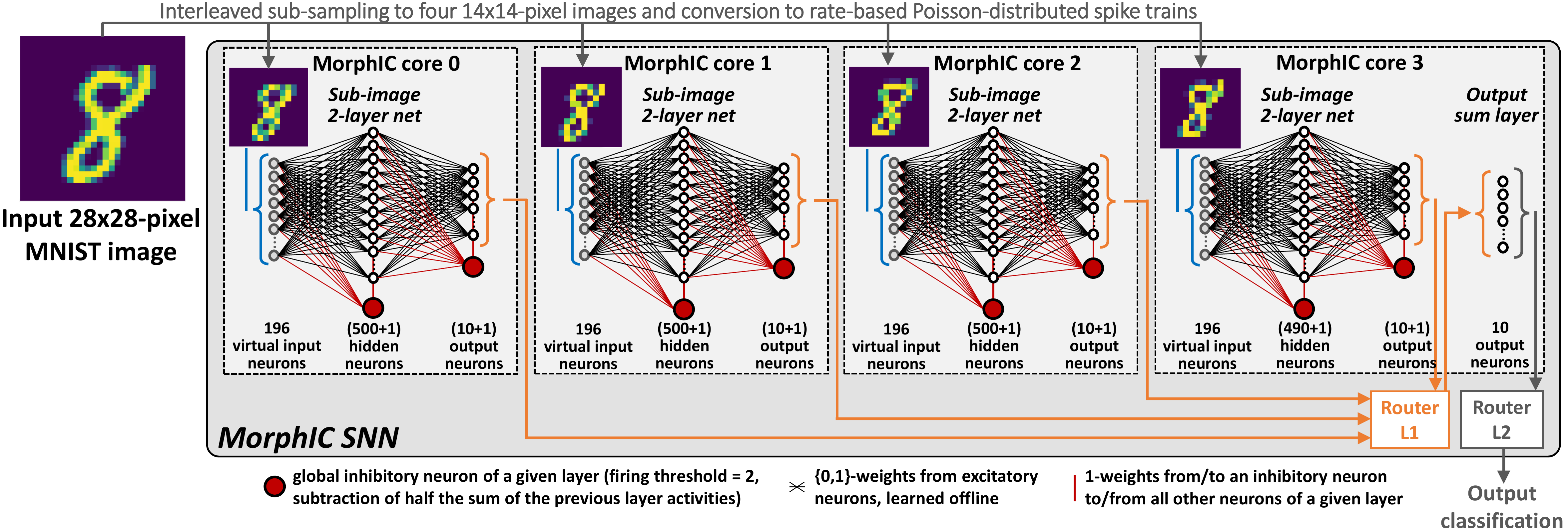}
\caption{MNIST classification setup. Input images are split with interleaved sub-sampling into four independent 14$\times$14 images. The sub-image pixels are converted to rate-based Poisson-distributed spike trains and sent to four one-hidden-layer fully-connected networks resulting from Adam-based quantization-aware training in Keras following~\mbox{\hspace{-0.1mm}\cite{Courbariaux16,Moons17}}. Layer-wise inhibitory neurons are used to compensate for rescaling of synaptic weights trained with $-1$ and $+1$ values in Keras to values of $0$ and $1$ in MorphIC. Average-pooling the core activities into a global output sum layer leads to a 97.8-\% classification accuracy. L1 (resp. L2) connectivity carries inferences from sub-classifiers (resp. combined sub-classifiers), illustrating that the level of the encoded features increases with the connectivity hierarchy.}%
\label{fig_mnist}
\end{figure*}

\section{Measurements and Benchmarking Results} \label{sec_results}

MorphIC was prototyped in the UMC 8-metal 65-nm low-power~(LP) CMOS process. A chip microphotograph is presented in Fig.~\ref{fig_chip}, while specifications and measurement results are provided in Table~\ref{table_specs}. A detailed area breakdown is provided in Table~\ref{table_area}. As derived in~\cite{Frenkel19}, the power consumption $P$ of time-multiplexed digital SNN architectures can be modeled by

\begin{equation}\label{eq_power}
P = P_\text{leak} + P_\text{idle}\times f_\text{clk} + E_\text{SOP}\times r_\text{SOP},
\end{equation}

\noindent where $P_\text{leak}$ is the leakage power, $P_\text{idle}$ is the idle power (i.e.~active clock, without network activity), $E_\text{SOP}$ is the energy per synaptic operation (SOP), $f_\text{clk}$ is the clock frequency and $r_\text{SOP}$ is the SOP processing rate. $E_\text{SOP}$ is an incremental definition of the energy per SOP as it does not include contributions from leakage and idle power. For example, based on Table~\ref{table_specs}, MorphIC consumes a total energy of 51pJ per SOP at 0.8V when including the leakage and idle power contributions at maximum $f_\text{clk}$ and $r_\text{SOP}$ (i.e. 55MHz and 110MSOP/s using all cores, each SOP taking two clock cycles as shown in Fig.~\ref{fig_clock_diagram}).

Offline learning performance with quantization-aware training can be demonstrated with the MNIST dataset of handwritten digits~\cite{LeCun98}. Using the four cores of MorphIC and all the available neuron resources with the network topology shown in Fig.~\ref{fig_mnist}, an accuracy of 97.8\% is reached using conventional rate-based coding (i.e. the spike frequency of a neuron encodes its output value). As the synaptic weights trained in Keras have $-$1 and $+$1 values while the MorphIC synapses have 0 and 1 values instead, it is necessary to compensate for the asymmetric weight distribution of MorphIC. This compensation can be made in a layer by subtracting half the sum of its inputs, which can be achieved by layer-wise inhibitory neurons connected with weight 1 to all the layer inputs and having a firing threshold of 2, as shown in Fig.~\ref{fig_mnist}. As the rate code is inefficient in its spike use, it results in a high energy per classification of 205$\mu$J at 0.8V and 55MHz. It has been shown in~\cite{Frenkel19} that the rank order code (i.e. values are encoded in the order in which neurons spike) is a simple yet much more efficient coding strategy than rate coding. The inferred class can be retrieved from the neuron in the output layer that spikes first. Using rank order coding, MorphIC consumes 5.45mW for 250 classifications per second at 0.8V and 55MHz, which allows reaching a 10-fold energy improvement down to 21.8$\mu$J per classification, at the expense of a drop of 1.9\% in accuracy. The energy-accuracy tradeoff of MorphIC will be discussed and compared to the state of the art in Section~\ref{ssec_ppa}.

\begin{figure*}[!t]
\centering
\noindent\includegraphics[width=0.835\textwidth]{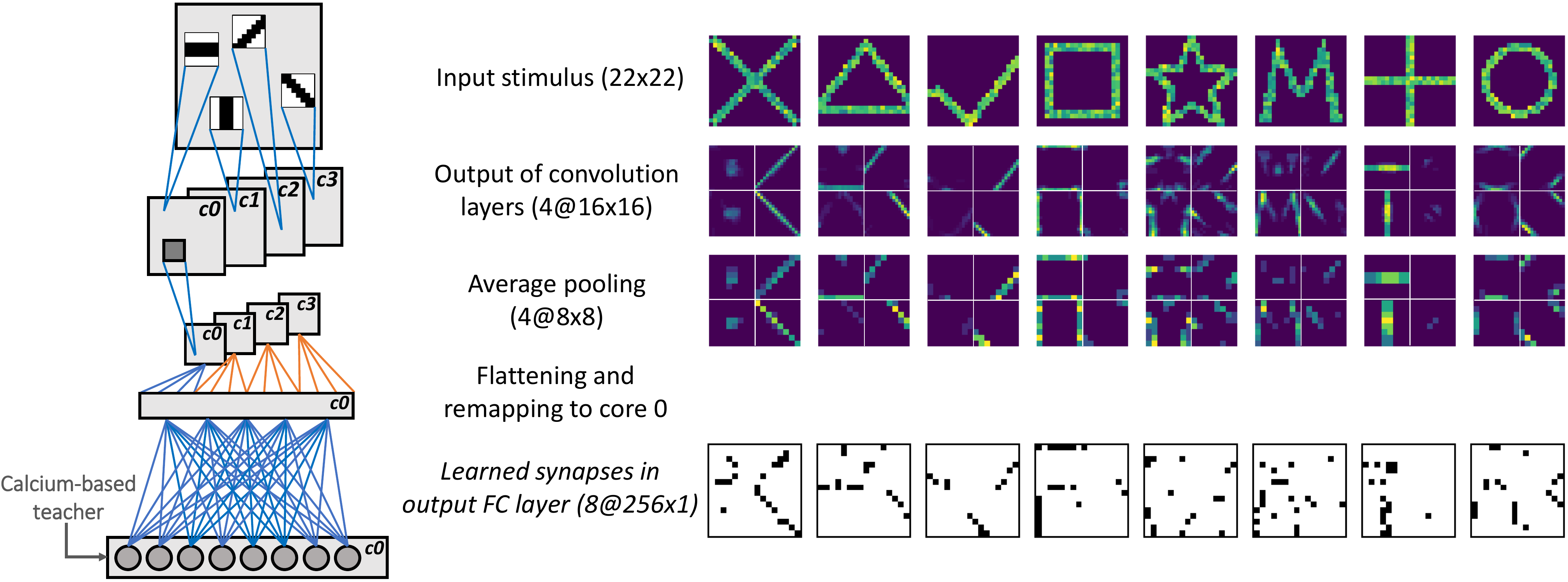}
\caption{MorphIC implementation of the 8-pattern CNN classification benchmark proposed in~\cite{Indiveri15}. Left: the CNN architecture consists of four 7$\times$7 line detection kernels at 0$^\circ$, 45$^\circ$, 90$^\circ$ and 135$^\circ$, followed by average pooling layers and a fully-connected (FC) 8-neuron output layer. Right: the input pattern activities, neuron activations of convolution and pooling layers and synaptic weights learned online with S-SDSP are illustrated. The test set consists of 100 different Poisson realizations of each input pattern, all 800 test samples are correctly classified.\vspace*{-0.5mm}}%
\label{fig_cnn}
\end{figure*}

S-SDSP online learning is demonstrated in Fig.~\ref{fig_cnn}, we reproduced the benchmark that was proposed in~\cite{Indiveri15} for an analog SDSP implementation. Eight patterns are classified by a spiking CNN. Each MorphIC core implements a fixed-weight convolutional layer with a line detection kernel followed by an average pooling layer. The pooling layers from cores 1 to 3 are then mapped back to core 0 through L1 connectivity so as to form a single flattened layer. The flattened layer is connected with plastic weights to an 8-neuron fully-connected~(FC) output layer in core 0. The resulting weights allow correctly discriminating all test samples in a test set consisting of 100 different Poisson realizations of each input pattern. 

\section{Discussion} \label{sec_disc}

Comparison with the state of the art can be carried out along several axes, all of which lead to guidelines for future work. Section~\ref{ssec_hierout} compares the proposed hierarchical event routing infrastructures with previously-proposed approaches. Section~\ref{ssec_ssdsp} discusses the implementation strategy of the proposed S-SDSP learning rule. Section~\ref{ssec_ppa} analyzes the area-accuracy and energy-accuracy tradeoffs on the MNIST dataset. Finally, Section~\ref{ssec_bnn} compares MorphIC with the three binary-weight SNNs proposed to date: TrueNorth~\cite{Akopyan15}, Loihi~\cite{Davies18} and the chip from Seo~\textit{et~al.}~\cite{Seo11}.

\subsection{Hierarchical event routing} \label{ssec_hierout}

A key feature of MorphIC is that only 27 bits per neuron are required to entirely define the connectivity of a multi-chip network, while the routers at all the hierarchy levels are memory-less beyond simple event buffering (i.e. no dedicated access to a stored mapping table). This contrasts with Neurogrid~\cite{Benjamin14} and HiAER~\cite{Park17}, which achieve low-cost large-scale routing at the expense of requiring external mapping table storage, thus indirectly inducing high resource and power overheads. SpiNNaker~\cite{Painkras13,Navaridas09} embeds the largest-scale multicast connectivity infrastructure proposed so far. It avoids external accesses but requires local mapping tables inside the routers to store all connectivity information for source-based 2-D triangular toroidal mesh-based routing. The same holds for DYNAPs~\cite{Moradi18}, which also avoids external accesses but still requires local 2.5-kB SRAM storage in the R1 router of each 256-neuron core, beyond a CAM-based 640-bit tag storage per neuron that allows reaching a high fan-out of 4k at the expense of density. Moreover, in DYNAPs, the synaptic connections are defined within the connectivity infrastructure, as opposed to MorphIC which has independent online-learning synapses on the top of the connectivity infrastructure. Therefore, none of Neurogrid, HiAER, SpiNNaker or DYNAPs offer low-cost memory-less routers throughout the hierarchy. The routing infrastructure of TrueNorth~\cite{Akopyan15} holds similar advantages to the one of MorphIC: the routers are also memory-less and require neither external nor internal storage. However, there are only two levels in the TrueNorth hierarchy: the 64k-synapse 256-neuron crossbar cores and the large-scale mesh-based routers. While an arbitrary number of neurons can map to the same axons\footnote{Axons are defined as inputs in the TrueNorth terminology, each axon connecting to 256 neurons through 256 synapses in a local crossbar array.}, thus reaching arbitrary fan-ins with 256 shared synapses, the neuron fan-out of TrueNorth is limited to a maximum value of 256+256. In MorphIC, the addition of an intermediate L1 star-based router brings a two-fold advantage over TrueNorth: it divides by two the number of L2 hops between any two cores and allows extending the neuron fan-out. Finally, the hierarchical routing infrastructure of Loihi~\cite{Davies18} differs strongly from all other previously-proposed approaches and, as DYNAPs, does not rely on a crossbar operation at the lowest level of the hierarchy. Instead, it is highly configurable in order to adapt to the target application: 120-kB SRAM memories are used to store the entire synaptic fan-in state of each core, which allows trading off the number of synapses with flexibility in the connectivity patterns. The high-level routing infrastructure of Loihi relies on a unicast mesh-based NoC: in order to process neuron spike multicasting with up to 4096 output axons, one packet needs to be generated for each destination. Therefore, Loihi achieves multicasting at the expense of high router load overhead. While the three-level connectivity infrastructure of MorphIC is less flexible, multicasting is only handled at the L1 level and a clear hierarchy segmentation avoids overloading higher-level routers. It ensures that the higher the hierarchy, the sparser the events and the higher the level of the encoded features, which leads to high efficiency if the hierarchical nature of the event routing infrastructure matches the intrinsic hierarchy in the data representation of a given task. In the case of MorphIC, as shown in Section~\ref{sec_results}, the chosen hierarchical routing infrastructure is ideal to process tasks of the complexity of MNIST in a single chip. The cores implement four weak classifiers that carry out inference on independent sub-images, star-based L1 routing allows all cores to be at equidistance (as opposed to mesh-based routing) to combine weak classifications in a single core, while L2 routing is used to transmit sparse spikes encoding the inference on the full MNIST image.%

As the hierarchical event routing fabric of MorphIC strongly relies on all-to-all crossbar operation~(Fig.~\ref{fig_noc_illus}), the proposed connectivity infrastructure is ideal to explore fully-connected and recurrent network configurations. Though convolutional layers can also be implemented using crossbars, the absence of efficient weight reuse implies copying the kernels in the dendritic tree of each output neuron. As the receptive field of each neuron in a convolutional layer usually consists of only a few inputs, there is a poor utilization of the synaptic fan-in resources while the time-multiplexed controller goes through a majority of dummy SOPs with zero weights~(Fig.~\ref{fig_clock_diagram}). The latter aspect can be mitigated by a controller update, similarly to the crossbar optimization with start and end addresses proposed in Section~\ref{ssec_ppa} for fully-connected layers.%

Finally, it is worth noting that all the routers in the aforementioned approaches operate asynchronously, except in HiAER~\cite{Park17} and MorphIC. In MorphIC, the choice of clocked operation for the routers allows for a straightforward design at the expense of efficiency. Indeed, the timing critical path being located at the core level, using one global clock common to both the cores and the routers unnecessarily limits the bandwidth of the latter which, as they are memory-less, could operate at much higher speed. Asynchronous router design would alleviate this problem. In order to avoid the design time and complexity overhead of asynchronous digital design, another option would be to locally generate a high-speed clock directly in the L1 and L2 routers, for example with a local ring oscillator that is enabled only when packets await routing.

\begin{figure*}[!t]
\centering
\vspace*{-2mm}
\noindent\includegraphics[width=0.99\textwidth]{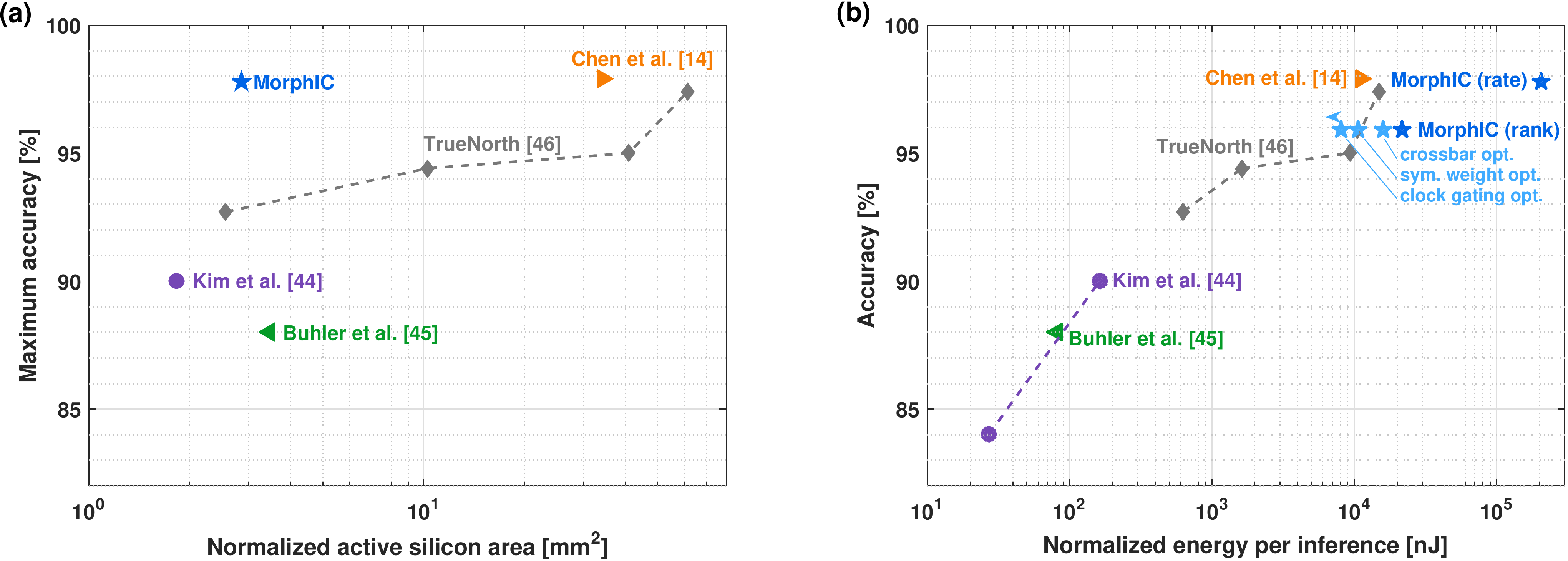}
\caption{Analysis of tradeoffs between accuracy, area and energy per classification on the MNIST dataset. (a) Area-accuracy tradeoff. Silicon area (excluding pads) has been  normalized to a 65-nm technology node using the node factor (e.g., a (65/40)$^2$-fold increase for normalizing 40nm to 65nm), except for the 10-nm FinFET node from Chen \textit{et~al.}~\cite{Chen18} where data from~\cite{Mistry17} was used for normalization. The TrueNorth area varies as Esser~\textit{et~al.} used different numbers of cores for their experiments (5, 20, 80 and 120 cores, in the order of increasing accuracy)~\cite{Esser15}. A 1920-core configuration is also reported in~\cite{Esser15}, leading to a 99.42-\% accuracy on MNIST with TrueNorth, a record for SNNs. However, as this configuration would lead to a normalized area of 980mm$^2$, we only included TrueNorth configurations whose scale are comparable with previously-proposed SNNs. MorphIC shows an order-of-magnitude improvement compared to previously-proposed approaches. (b) Energy-accuracy tradeoff. Energy has been  normalized to a 65-nm technology node using the node factor (e.g., a (65/40)-fold increase for normalizing 40nm to 65nm). MorphIC demonstrates an energy-accuracy tradeoff close to TrueNorth and the SNN from Chen~\textit{et~al.}~\cite{Chen18} using rank order coding, compared to an unfavorable tradeoff with rate coding (dark blue). In light blue are shown power savings that could be achieved with three simple optimizations in the chip architecture or implementation, estimated from the power model of Eq.~(\ref{eq_power}).\vspace*{-1mm}}
\label{fig_ppa}
\end{figure*}

\subsection{S-SDSP online learning}  \label{ssec_ssdsp}

\begin{table}[!t]
\caption{Benchmark summary for silicon implementations of bottom-up STDP- and SDSP-based learning rules.}
\label{tab_bench}
\centering
\resizebox{0.98\columnwidth}{!}{
\begin{tabular}{lccc}
\toprule%
Chip(s) & Implementation & Learning rule & Benchmark \\
\midrule%
BrainScaleS~\cite{Schemmel10} & Analog & STDP & -- \\
DYNAPs + ROLLS~\cite{Indiveri15} & Analog & SDSP & 8-pattern classif. \\
Mayr \textit{et al.}~\cite{Mayr16} & Analog & SDSP & -- \\
Seo \textit{et al.}~\cite{Seo11} & Digital & 1-bit S-STDP & 2-pattern recall \\
Chen \textit{et al.}~\cite{Chen18} & Digital & 7-bit STDP & Denoising \\
Loihi~\cite{Davies18} & Digital & STDP-based & Pre-processed MNIST \\
ODIN~\cite{Frenkel19} & Digital & 4-bit SDSP & 16$\times$16 deskewed MNIST \\
MorphIC [This work] & Digital & 1-bit S-SDSP & 8-pattern classif. \\
\bottomrule%
\end{tabular}}%
\end{table}

The development of silicon implementations for bio-inspired learning rules such as STDP and SDSP is an inherently \textit{bottom-up} process: the first step lies in neuroscience experimentation to understand biological phenomena, the second step lies in the development of efficient analog or digital circuits that emulate neuroscience observations, the third step consists in finding a suitable application. Therefore, while bottom-up approaches lead to efficient silicon implementations of bio-inspired plasticity models and are ideal for the design of experimentation platforms, bridging the gap from local brain-inspired learning toward more complex real-life applications is difficult. To the best of our knowledge, only simplified benchmarks have been used so far to demonstrate silicon implementations of STDP, SDSP or their variations. A summary is provided in Table~\ref{tab_bench}: no STDP- or SDSP-based learning rule has yet been successfully applied \textit{in silico} to at least the full MNIST dataset without any pre-processing step. The S-SDSP learning rule we propose for MorphIC allows reaching the highest density of online-learning synapses and has successfully been demonstrated on the 8-pattern benchmark from~\cite{Indiveri15}. However, as it also follows from a bottom-up design approach, scaling S-SDSP to more complex tasks is not straightforward as it would require going beyond single-layer training. Further research is required to leverage brain-inspired local plasticity primitives with multi-layer networks for online learning on real-world tasks, as highlighted by the recent S-STDP study~\mbox{by Yousefzadeh \textit{et~al.}~\cite{Yousefzadeh18}.}%

On the other hand, \textit{top-down} approaches start from the applicative problems (e.g., image recognition), where the successful backpropagation of errors algorithm~\cite{Rumelhart86} has already been applied to specific datasets such as MNIST, CIFAR-10 or ImageNet. Such approaches then attempt to design variations of the backpropagation algorithm that are more in line with brain observation, such as moving data representation to spiking or dropping the requirement for symmetric weights~\cite{Neftci17,Guerguiev17}. Silicon implementations would come as a last step. In order to overcome the challenges of bottom-up approaches, the development of new multi-layer spike-based learning rules following top-down approaches has gained growing interest in the recent years (e.g.,~\cite{Zheng17,Mostafa17,Neftci17,Zenke18}). Further research is yet required to realize efficient silicon implementations of such learning rules and to make them both compatible with an online-learning setup and able to leverage weight quantization down to binary or ternary resolutions.

Finally, regarding the synapse implementation, we showed in Section~\ref{sec_ssdsp} that our S-SDSP design is compatible with a standard single-port foundry SRAM, which holds a strong advantage in design time and density over custom-SRAM-based designs, such as in~\cite{Seo11}. Our foundry-SRAM-based S-SDSP approach therefore allows leveraging high-density integration of binary plastic synapses. There are two other main trends for synapse implementation. First, the capacitor-based approach proposed in~\cite{Qiao15} for the subthreshold analog 0.18-$\mu$m ROLLS chip allows emulating SDSP dynamics with biological time constants and a resolution of a few bits, at the expense of synaptic mismatch and a critical silicon footprint for the pF-range capacitor inside each synapse. Therefore, technology scaling pushes recent subthreshold analog developments to move synaptic weights to TCAM and SRAM memories~\cite{DeSalvo18}. Second, non-volatile memories for crossbar implementations leveraging in-memory computation with novel technologies are currently being actively explored. On the one hand, a flash-based approach with STDP plasticity was successfully prototyped in 0.35$\mu$m CMOS in~\cite{Brink13}, however embedded flash memory is difficult to scale beyond 40nm and requires high programming voltages. On the other hand, memristors promise new density records and recent work (e.g.,~\cite{Payvand18}) shows how the memristor characteristics can be used to emulate biological synapses and to implement stochastic learning, but high-yield co-integration with CMOS has yet to be demonstrated. For both flash- and memristor-based approaches, the aspects linked to synaptic resolution control, mismatch and fabrication costs will have to be alleviated. It therefore appears that the foundry-SRAM-based strategy that we propose is currently a sound strategy for an efficient low-cost synapse array design.

\subsection{Tradeoff analysis of energy, area and accuracy}  \label{ssec_ppa}

An analysis of the energy, area and accuracy tradeoffs is shown in Fig.~\ref{fig_ppa}, where MorphIC is compared to other SNNs that have been demonstrated on the full 28$\times$28 MNIST dataset with no pre-processing beyond conversion of pixel values to spikes: the SNNs from Chen \textit{et~al.}~\cite{Chen18}, from Kim~\textit{et~al.}~\cite{Kim15}, from Buhler~\textit{et~al.}~\cite{Buhler17} and TrueNorth, which was benchmarked on MNIST in~\cite{Esser15}. In order to carry out comparison in a one-to-one basis, all area and energy numbers have been normalized to a 65-nm technology node. While we keep this comparison focused on SNNs, there is also a large body of work for efficient MNIST-proven frame-based neural network accelerators: we refer the reader to~\cite{Whatmough17} for a partial review and to~\cite{Chen19} for a recent example.

\begin{table}[!t]
\caption{Summary of simple architectural and implementation improvements to MorphIC that would allow to further reduce the energy per inference on MNIST at 0.8V and 55MHz.}
\label{tab_opt}
\centering
\resizebox{0.98\columnwidth}{!}{
\begin{tabular}{lcccc}
\toprule%
Optimization & $E_\text{leak}$ & $E_\text{idle}$ & $E_\text{SOPs}$ & $E_\text{infer}$ \\
\midrule%
MorphIC, rank order & \multirow{2}{*}{0.18$\mu$J} & \multirow{2}{*}{9.24$\mu$J} & \multirow{2}{*}{12.4$\mu$J} & \multirow{2}{*}{21.8$\mu$J} \\
code, no improvement & & & & \\
+ crossbar opt. & 0.13$\mu$J & 6.72$\mu$J & 9.15$\mu$J & 16.0$\mu$J \\
+ sym. weight opt. & 0.09$\mu$J & 4.48$\mu$J & 6.10$\mu$J & 10.7$\mu$J \\
+ clock gating opt. & 0.09$\mu$J & 2.02$\mu$J & 6.10$\mu$J & 8.2$\mu$J \\
\bottomrule%
\end{tabular}}%
\end{table}

The area-accuracy tradeoff is shown in Fig.~\ref{fig_ppa}(a): MorphIC achieves an order-of-magnitude improvement compared to previously-proposed SNNs. The energy-accuracy tradeoff is shown in Fig.~\ref{fig_ppa}(b). While rate coding allows reaching the highest accuracy for MorphIC, the associated power inefficiency is clearly illustrated. Rank order coding (Section~\ref{sec_results}) allows reaching an energy-accuracy tradeoff that comes close to TrueNorth and the SNN from Chen \textit{et~al.}~\cite{Chen18}. Following the MorphIC power model from Eq.~(\ref{eq_power}), we can break down the rank-order energy per inference $E_\text{infer}$ of 21.8$\mu$J at 0.8V and 55MHz as follows: $E_\text{infer} = E_\text{leak} + E_\text{idle} + E_\text{SOPs}$. Three simple improvements to the MorphIC architecture or implementation would allow to reduce these contributions, as shown in Fig.~\ref{fig_ppa}(b) and detailed in Table~\ref{tab_opt}. First, while crossbar operation is highly-efficient for the implementation of fully-connected layers, systematic processing of all neurons in the array for each input spike can lead to a lot of dummy operations (e.g., as shown in Fig.~\ref{fig_mnist}, a neuron from the hidden layer should lead to only 10+1 SOPs toward the output layer neurons, not 512). This could be improved by adding only two 9-bit parameters per neuron that define the start and end indices of the destination neurons. Adding this architectural improvement to MorphIC would reduce the number of SOPs to be processed and thus the time per inference, bringing $E_\text{infer}$ down to 16.0$\mu$J per inference. Second, as shown in Fig.~\ref{fig_mnist}, inhibitory neurons have been added to the hidden and output layers to compensate for rescaling of the synaptic weights trained offline with $-1$ and $+1$ values to $0$ and $1$ values in MorphIC. However this comes at the expense of a 50-\% increase in SOP activity. In combination with the previous improvement, allowing the MorphIC binary weights to be interpreted as $-1$ and $+1$ values instead of $0$ and $1$ would allow to further reduce $E_\text{infer}$ to 10.7$\mu$J per inference. Finally, as clock gating has only been inserted automatically by the synthesis tool, simple architectural clock gating could be applied (i)~to the parameter banks after initial chip configuration and (ii) to S-SDSP up/down registers (Fig.~\ref{fig_morphic_core}) during inference. This would lead to an idle power reduction by 55\%, further optimizing $E_\text{infer}$ down to a value of 8.2$\mu$J per inference.

Therefore, MorphIC demonstrates an order-of-magnitude improvement in the area-accuracy tradeoff on the MNIST classification task, while keeping an energy-accuracy tradeoff comparable to TrueNorth and the SNN from Chen \textit{et~al.}~\cite{Chen18}. With a low leakage of only 45$\mu$W at 0.8V to ensure full retention of the neuronal and synaptic data, MorphIC is ideally-suited for always-on event-driven processing.

\subsection{Comparison with previously-proposed binary SNNs}  \label{ssec_bnn}

A comparison of MorphIC with the three previously-proposed binary SNNs is provided in Table~\ref{table_SoA}. As TrueNorth embeds static weights and Loihi has a programmable learning engine but does not demonstrate online learning with a binary-weight configuration, MorphIC and the chip from Seo~\textit{et~al.}~\cite{Seo11} are the only ones to demonstrate embedded online learning on binary weights. The high-density claim of binary-weight S-SDSP online learning is demonstrated with an order-of-magnitude advantage compared to the S-STDP rule from Seo~\textit{et~al.}~\cite{Seo11}. This point is further emphasized when considering process normalization to 65nm, illustrating record densities for MorphIC. Regarding power, MorphIC has an energy per SOP similar to the other binary SNNs despite using a less advanced CMOS process.

\begin{table}
\setlength\tabcolsep{4pt} %
\caption{Comparison of binary SNN processor chips.}%
\label{table_SoA}
\renewcommand{\arraystretch}{1.07}
\centering
\resizebox{\columnwidth}{!}{
\begin{tabular}{lcccc}
\toprule%

Reference &  Seo \textit{et al.}~\cite{Seo11} & TrueNorth~\cite{Akopyan15} & Loihi~\cite{Davies18} & \textbf{This work} \\\midrule

Technology &  45nm SOI & 28nm LP & 14nm FinFET & 65nm LP \\

Area (excl. pads) & 0.8mm$^2$ & 389mm$^2$ & 51.8mm$^2$ & 2.86mm$^2$ \\

\# cores & 1 & 4096 & 128 & 4 \\

\# neurons / core & 256 & 256 & max. 1024 & 512 \\

\# synapses  / core & 64k & 64k & 114k to 1M & 528k  \\

Synaptic width & 1-bit & 1-bit & 9- to 1-bit & 1-bit \\

On-chip learning & S-STDP & -- & Programmable & S-SDSP \\

Flexibility \begin{tabular}{@{}c@{}}~routing\\~learning\end{tabular} & \begin{tabular}{@{}c@{}}Low\\Low\end{tabular} & \begin{tabular}{@{}c@{}}Medium\\--\end{tabular} & \begin{tabular}{@{}c@{}}High\\High\end{tabular} & \begin{tabular}{@{}c@{}}Medium\\Low\end{tabular} \\

Neuron density$^\dagger$ \begin{tabular}{@{}c@{}}raw\\norm.\end{tabular} & \begin{tabular}{@{}c@{}}320\\153\end{tabular} & \begin{tabular}{@{}c@{}}2.6k\\494\end{tabular} & \begin{tabular}{@{}c@{}}max. 2.5k\\max. 190\end{tabular} & \begin{tabular}{@{}c@{}}716\\716\end{tabular} \\%

Synapse density$^\dagger$ \begin{tabular}{@{}c@{}}raw\\norm.\end{tabular} & \begin{tabular}{@{}c@{}}80k\\38.3k\end{tabular} & \begin{tabular}{@{}c@{}}674k\\125k\end{tabular} & \begin{tabular}{@{}c@{}}282k to 2.5M\\21k to 190k\end{tabular} & \begin{tabular}{@{}c@{}}738k\\738k\end{tabular} \\ %

Incremental energy/SOP & N/A & N/A & ($>$23.6pJ at 0.75V)$^*$ & 30pJ at 0.8V \\

Total energy/SOP & N/A & 26pJ at 0.775V & N/A & 51pJ at 0.8V \\

\bottomrule
\end{tabular}}
\vspace*{0.5mm}

\scriptsize

$^*$ Simulation results, excluding the cost of neuron and learning engine updates.
\vspace*{0.5mm}

\hspace*{-1.8mm}$^\dagger$ Neuron and synapse densities are obtained by dividing the total number of neurons or~synapses by the chip area, excluding pads. As the raw density performance is strongly dependent on the selected technology node, values normalized to a 65-nm node are provided. Normalization is carried out by using the node factor, except for Loihi where we used data from~\cite{Mistry17} for 14nm FinFET normalization to bulk 65nm.

\end{table}

\section{Conclusion} \label{sec_conclusion}

In this paper, we presented the MorphIC quad-core spiking neuromorphic processor to leverage binary weights with online-learning SNNs. Using the proposed stochastic spike-dependent synaptic plasticity (S-SDSP) learning rule, we demonstrated this claim with a density of 738k synapses per mm$^2$ in 65nm CMOS. MorphIC shows order-of-magnitude improvements both in the area-accuracy tradeoff on MNIST compared to other SNNs and in density compared to the only previously-proposed binary SNN with demonstrated online learning from Seo~\textit{et~al.}~\cite{Seo11}. MorphIC also integrates a low-cost hierarchical routing fabric with low-memory connectivity for large-scale chip interconnection, where distance information allows modulating the synaptic update probabilities, in accordance with small-world brain network modeling.

\section*{Acknowledgment}

The authors would like to thank Europractice and its First User Stimulation Program for chip prototyping. C.~Frenkel is with Universit\'e catholique de Louvain as a Research Fellow from the National Foundation for Scientific Research (FNRS) of Belgium.

\begin{IEEEbiography}
	[{\includegraphics[width=1in,height=1.25in,clip,keepaspectratio]{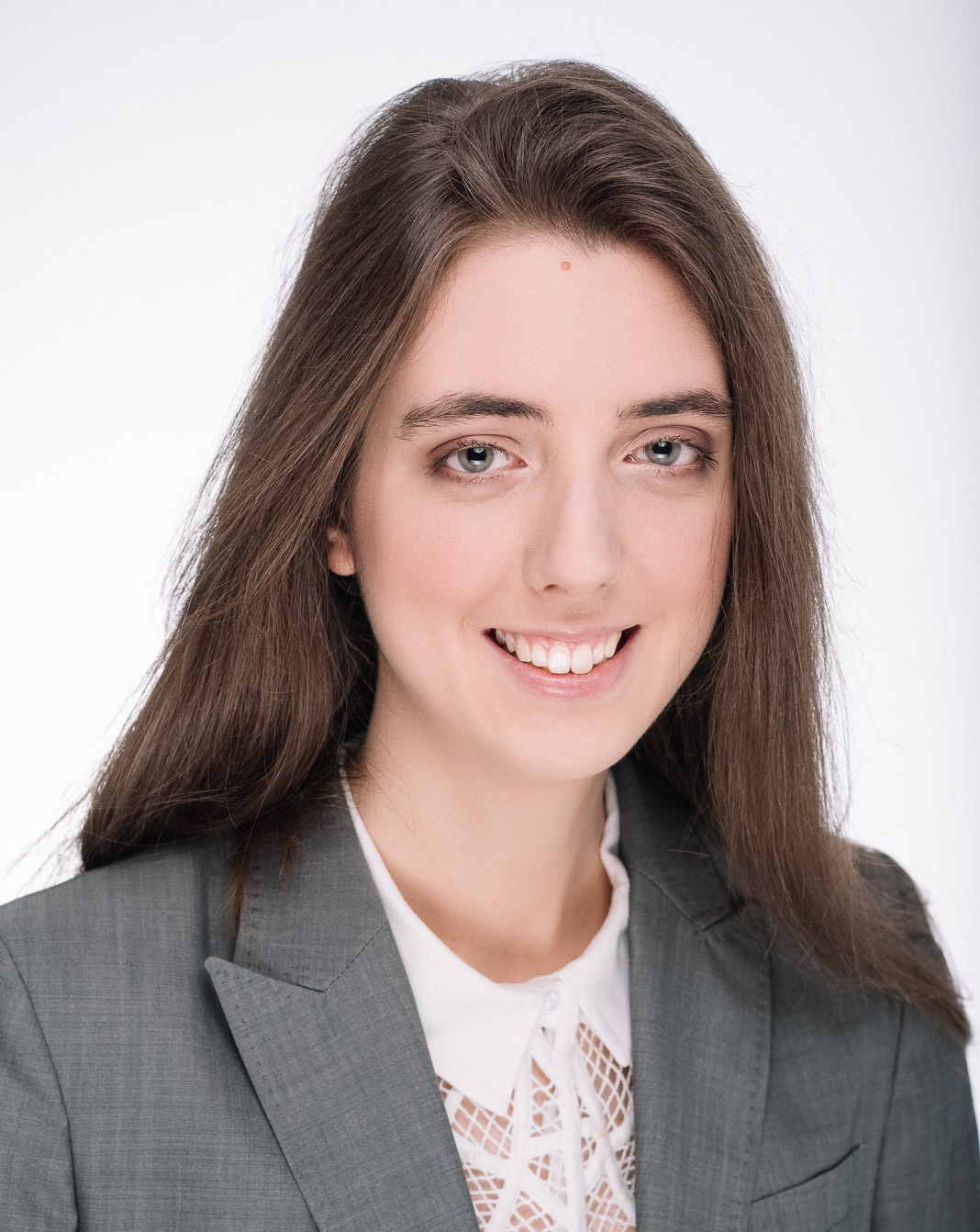}}]{Charlotte Frenkel}
(S'15) received the M.Sc. degree (\textit{summa cum laude}) in Electromechanical Engineering from Universit\'e catholique de Louvain (UCLouvain), Louvain-la-Neuve, Belgium, in 2015. She is currently working toward the Ph.D. degree as a Research Fellow of the National Foundation for Scientific Research (FNRS) of Belgium, under the supervision of Prof. D. Bol and Prof. J.-D. Legat.

Her current research focuses on the design of low-power and high-density neuromorphic circuits as efficient non-von-Neumann architectures for real-time recognition and online learning.

Ms. Frenkel serves as a TPC member for the IEEE MCSoC conference and as a reviewer for the IEEE Trans. on Neural Networks and Learning Systems, Trans. on Cognitive and Developmental Systems journals and for the IEEE ISCAS, BioCAS, S3S conferences.
\end{IEEEbiography}

\vskip -22pt plus -1fil

\begin{IEEEbiography}
	[{\includegraphics[width=1in,height=1.25in,clip,keepaspectratio]{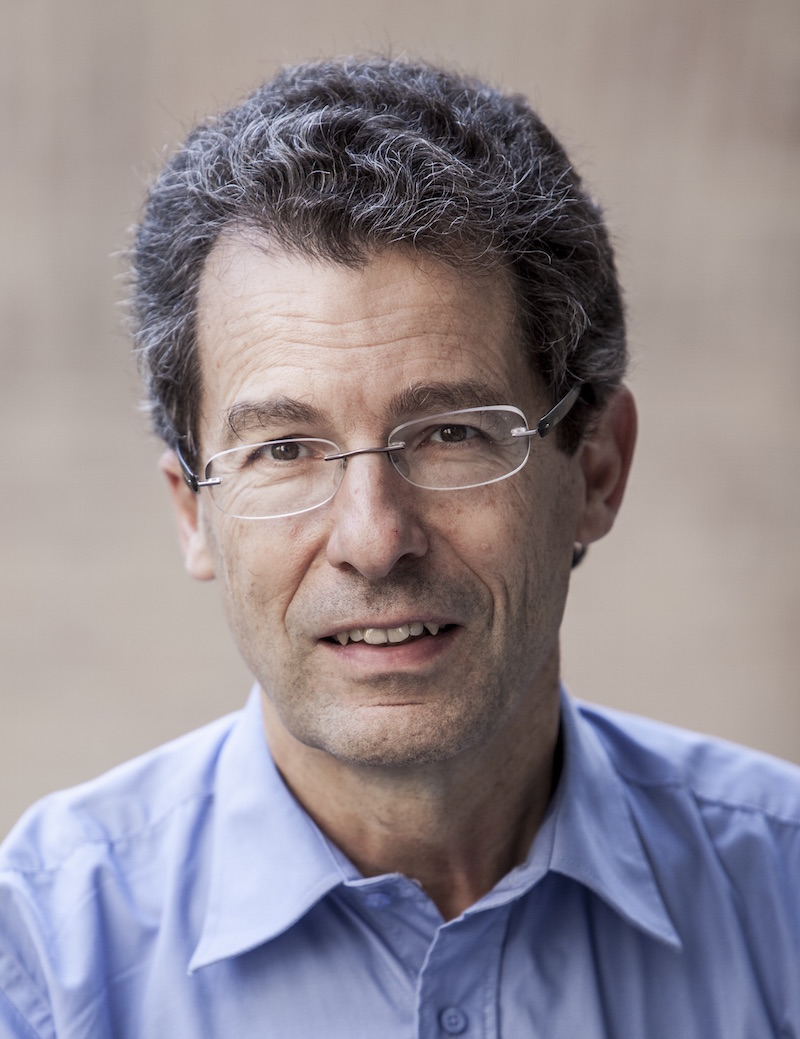}}]{Jean-Didier Legat}
(S'79-M'85-SM'17) received his engineering and PhD degrees in microelectronics from the Universit\'e catholique de Louvain, Louvain-la-Neuve, Belgium in 1981 and 1987, respectively.

From 1987 to 1990, he was with Image Recognition Integrated Systems (I.R.I.S.), a new company specialised in optical character recognition and automatic document processing. Jean-Didier Legat was co-founder and Vice-President of I.R.I.S. In October 1990, he came back to the UCLouvain Microelectronics Laboratory. He is presently full Professor. From 2003 to 2008, he was the Dean of the Louvain School of Engineering. Currently, he is Senior Advisor to the President for Technology Transfer and Head of the ICTEAM Research Institute. His current interests are processor architecture, low-power digital integrated circuits, real-time embedded systems, mixed-signal design and hardware-software codesign for reconfigurable systems. He has been an author or co-author of more than 200 publications in the field of microelectronics, low-power digital circuits, computer architecture, digital signal processing, computer vision and pattern recognition.
\end{IEEEbiography}

\vskip -22pt plus -1fil

\begin{IEEEbiography}
	[{\includegraphics[width=1in,height=1.25in,clip,keepaspectratio]{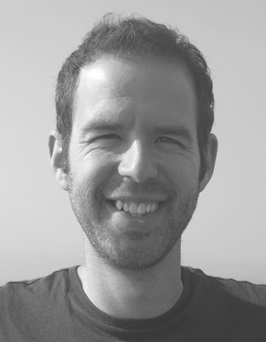}}]{David Bol}
(S'07-M'09-SM'18) received the M.Sc. degree in Electromechanical Engineering and the Ph.D degree in Engineering Science from Universit\'e catholique de Louvain (UCLouvain), Louvain-la-Neuve, Belgium in 2004 and 2008, respectively. In 2005, he was a visiting Ph.D student at the CNM National Centre for Microelectronics, Sevilla, Spain, in advanced logic design. In 2009, he was a postdoctoral researcher at intoPIX, Louvain-la-Neuve, Belgium, in low-power design for JPEG2000 image processing. In 2010, he was a visiting postdoctoral researcher at the UC Berkeley Laboratory for Manufacturing and Sustainability, Berkeley, CA, in life-cycle assessment of the semiconductor environmental impact. He is now an assistant professor at UCLouvain. In 2015, he participated to the creation of e-peas semiconductors, Louvain-la-Neuve, Belgium.

Prof. Bol leads with Prof. Denis Flandre the Electronic Circuits and Systems (ECS) research group focused on ultra-low-power design
of smart-sensor integrated circuits for the IoT and biomedical applications with a specific focus on environmental sustainability. His personal IC interests include computing, power management, sensing and wireless communications. He gives four M.Sc. courses
in Electrical Engineering at UCLouvain on digital, analog and mixed-signal integrated circuits and systems as well as sensors.

Prof. Bol has authored or co-authored more than 100 technical papers and conference contributions and holds three delivered patents. He (co-)received three Best Paper/Poster/Design Awards in IEEE conferences (ICCD 2008, SOI Conf. 2008, FTFC 2014). He also serves as an editor for MDPI J. Low-Power Electronics and Applications, as a TPC member of IEEE SubVt/S3S conference and as a reviewer for various journals and conferences such as IEEE J. of Solid-State Circuits, IEEE Trans. on VLSI Syst., IEEE Trans. on Circuits and Syst. I/II. Since 2008, he presented several invited papers and keynote tutorials in international conferences including a forum presentation at IEEE ISSCC 2018.%
\end{IEEEbiography}

\end{document}